\documentclass[sigconf]{acmart}

\usepackage{amsmath}
\usepackage{algorithm}
\usepackage{algorithmic}
\usepackage{booktabs}
\usepackage{graphicx}
\graphicspath{{../}{}}
\usepackage{xcolor}
\usepackage{multirow}
\usepackage{makecell}
\usepackage{stfloats}
\usepackage{afterpage}

\makeatletter
\def\@affiliationfont{\small\normalfont}
\makeatother

\copyrightyear{2026}
\acmYear{2026}
\setcopyright{cc}
\setcctype{by}
\acmConference[MM '26]{Proceedings of the 34th ACM International Conference on Multimedia}{November 10--14, 2026}{Rio de Janeiro, Brazil}
\acmBooktitle{Proceedings of the 34th ACM International Conference on Multimedia (MM '26), November 10--14, 2026, Rio de Janeiro, Brazil}
\acmISBN{979-8-4007-2213-4/2026/11}
\acmDOI{10.1145/3767308.3835233}

\acmSubmissionID{1875}

\AtBeginDocument{%
  }

\begin{document}
\title{DriveVLA-M0: Failure-Aware Memory Augmentation for Autonomous Driving}

\author{Zebin Xing}
\authornote{Equal contribution.}
\affiliation{%
  \institution{Institute of Automation, Chinese Academy of Sciences}
  \city{Beijing}
  \country{China}}
\email{xingzebin2024@ia.ac.cn}

\author{Yupeng Zheng}
\authornotemark[1]
\affiliation{%
  \institution{Institute of Automation, Chinese Academy of Sciences}
  \city{Beijing}
  \country{China}}
\email{zhengyupeng2022@ia.ac.cn}

\author{Qiang Chen}
\affiliation{%
  \institution{Institute of Automation, Chinese Academy of Sciences}
  \city{Beijing}
  \country{China}}
\email{jonathanchencasia@gmail.com}

\author{Linbo Wang}
\affiliation{%
  \institution{Institute of Automation, Chinese Academy of Sciences}
  \city{Beijing}
  \country{China}}
\email{wanglinbo2026@ia.ac.cn}

\author{Yichen Zhang}
\affiliation{%
  \institution{Institute of Automation, Chinese Academy of Sciences}
  \city{Beijing}
  \country{China}}
\email{zhangyichen2026@ia.ac.cn}

\author{Pengxuan Yang}
\affiliation{%
  \institution{Institute of Automation, Chinese Academy of Sciences}
  \city{Beijing}
  \country{China}}
\email{yangpengxuan20@mails.ucas.ac.cn}

\author{Junli Wang}
\affiliation{%
  \institution{Institute of Automation, Chinese Academy of Sciences}
  \city{Beijing}
  \country{China}}
\email{wangjunli2022@ia.ac.cn}

\author{Deheng Qian}
\affiliation{%
  \institution{Chongqing Chang'an Technology Co., Ltd.}
  \city{Chongqing}
  \country{China}}
\email{qiandeheng@qq.com}

\author{Xiaoqing Ye}
\affiliation{%
  \institution{Chongqing Chang'an Technology Co., Ltd.}
  \city{Chongqing}
  \country{China}}
\email{yxq@whu.edu.cn}

\author{Junyu Han}
\affiliation{%
  \institution{Chongqing Chang'an Technology Co., Ltd.}
  \city{Chongqing}
  \country{China}}
\email{pengjie.han@gmail.com}

\author{Yifeng Pan}
\affiliation{%
  \institution{Chongqing Chang'an Technology Co., Ltd.}
  \city{Chongqing}
  \country{China}}
\email{y.f.pan@hotmail.com}

\author{Qichao Zhang}
\authornote{Corresponding author.}
\affiliation{%
  \institution{Institute of Automation, Chinese Academy of Sciences}
  \city{Beijing}
  \country{China}}
\email{zhangqichao2014@ia.ac.cn}

\author{Dongbin Zhao}
\affiliation{%
  \institution{Institute of Automation, Chinese Academy of Sciences}
  \city{Beijing}
  \country{China}}
\email{dongbin.zhao@ia.ac.cn}

\renewcommand{\shortauthors}{Xing and Zheng et al.}


\begin{abstract}
Vision-Language-Action (VLA) models have recently emerged as a promising paradigm for end-to-end autonomous driving by enabling unified reasoning across perception, language, and planning.
However, existing approaches lack mechanisms to exploit past failures or adapt to distribution shifts, causing the model to persistently underperform on similar scenarios where it has previously failed.
In this paper, we propose DriveVLA-M0, a retrieval-augmented VLA with failure-aware latent memory. We construct a latent memory pool that stores failure cases along with their structure scene representations and expert trajectory labels, and design a dedicated Retrieve Model that decouples static road structure and dynamic agent interactions to enable structurally grounded retrieval. At inference time, retrieved cases are injected into the model via a lightweight decoupled LoRA-based test-time training (TTT) mechanism, allowing targeted and scenario-specific correction without modifying the backbone. 
Extensive experiments on NAVSIMv1 and NAVSIMv2 benchmark demonstrate that our approach consistently outperforms prior methods, achieving 94.1 PDMS on Navtest and 47.0 EPDMS on Navhard with only 26.44 ms TTT backward latency overhead. Furthermore, we show that DriveVLA-M0 scales effectively with additional memory, enabling training-free performance gains through memory expansion. The code is available at https://github.com/ZebinX/DriveVLA-M0.
\end{abstract}







\begin{teaserfigure}
  \centering
  \includegraphics[width=0.82\textwidth]{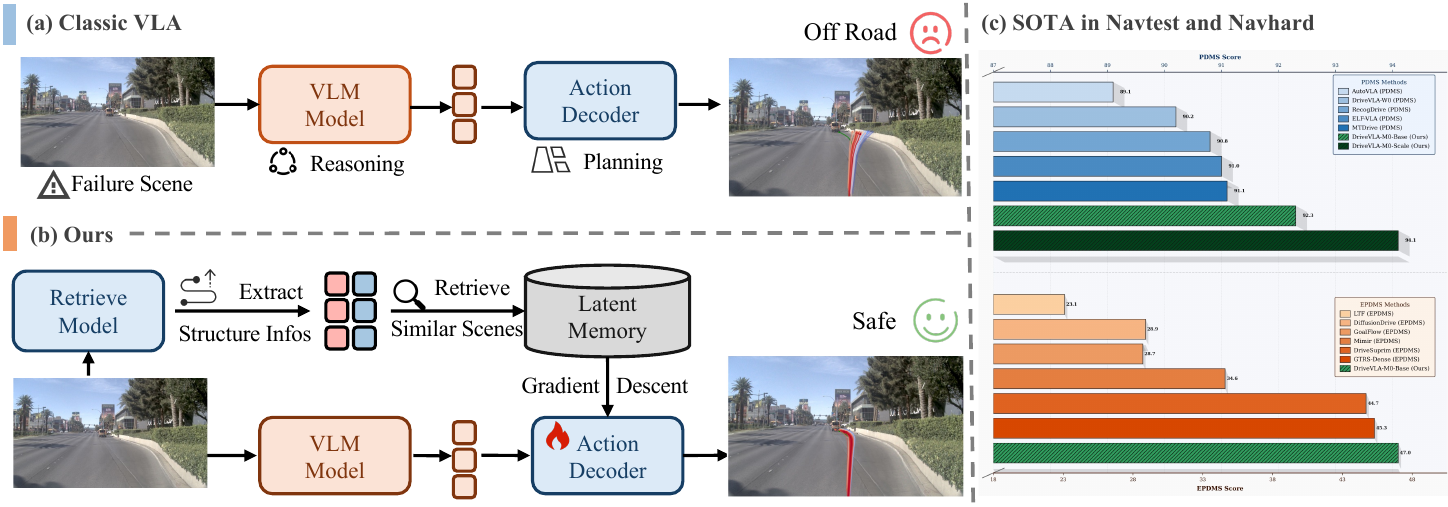}
 \caption{
  We propose DriveVLA-M0, a failure-aware memory augmentation VLA framework.
  (a) Classic VLA: VLMs perform scene reasoning in vision-language space and pass intermediate features to an Action Decoder for planning. The red trajectories indicate the model’s likely selections.
  (b) Ours: We retrieve structurally similar failure cases using Retrieved Model from a latent memory pool and inject them into the Action Decoder via LoRA-based test-time training for scenario-specific correction.
  (c) Performance: DriveVLA-M0 achieves SOTA on Navtest (NAVSIMv1, blue) and Navhard (NAVSIMv2, orange).
  }
  \label{fig:teaser}
\end{teaserfigure}

\maketitle


\section{Introduction}

Vision-language-action models in autonomous driving leverage large vision-language models (VLMs) to integrate semantic reasoning and high-level scene understanding into driving decisions, attracting significant interest for their potential in generalization and scalability. Early work, such as LMDrive~\cite{shao2024lmdrive} demonstrated the important role of vision-language models in following driving instructions. Subsequent studies~\cite{drivevlm, recogdrive, autovla}, show that reasoning mechanisms based on language models can substantially improve performance in complex driving scenarios.

Despite these promising results, existing VLA‑based driving models often repeat similar mistakes in similar scenarios, lacking the ability to directly associate the current situation with past failures. Studies~\cite{brown2005learned,botvinick2001conflict} on human physiology and behavior have shown that humans link ongoing situations with historical errors, predicting the likelihood of failure in the current context and adjusting their behavior accordingly. Motivated by this, we introduce an explicit latent memory that records failure cases, enabling associative retrieval and error likelihood assessment at test time, allowing the model to selectively adjust its planning based on prior failures.

In parallel, memory-augmented methods have shown strong potential across agent systems, robotics, and sequence modeling. Approaches such as MANTRA~\cite{mantra} and MemoNet~\cite{memonet} retrieve prototypical trajectories for motion prediction, while agent systems like JARVIS-1~\cite{jarvis1} and MemGen~\cite{memgen} leverage episodic memory for long-horizon decision-making. More recently, memory has also been introduced into VLA-style embodied agents, such as MemoryVLA~\cite{memoryvla} and EchoVLA~\cite{echovla}, primarily to improve long-horizon reasoning and task consistency.

Existing VLA-based methods in Embodied AI typically use intermediate vision-language features as retrieval keys for memory. However, unlike robotic manipulation, autonomous driving planning relies more heavily on two types of intrinsic information: (1) \textbf{dynamic information} (e.g., motion of surrounding agents), and (2) \textbf{scene structure information} (e.g., road topology). Using vision-language features directly as retrieval keys makes it difficult to effectively capture these two critical aspects, leading to retrieved cases that may differ significantly from the current scenario in terms of dynamics and structure—differences that are essential for robust decision-making and failure recovery.

To address this, we propose DriveVLA-M0~\ref{fig:teaser}, a framework that integrates failure-aware memory with test-time adaptation for VLA-based driving. Our key insight is that effective memory augmentation for autonomous driving requires two properties: (1) structural retrieval: retrieving cases based on scene-level similarity with respect to dynamic and physical structure rather than vision-language similarity; and (2) failure awareness: focusing on scenarios where the base model underperforms.

Our approach consists of two stages. In the offline stage (Memory Generation), we construct a latent memory pool by identifying failure scenarios via oracle simulation metrics and storing their intermediate representations, including scene embeddings, trajectory clusters, and expert signals. To enable effective retrieval, we train a dedicated Retrieve Model that explicitly captures static road structure and dynamic agent interactions using a dual-branch design. In the online stage (Inference with TTT), structurally similar cases are retrieved from memory and used to adapt the action decoder via a decoupled LoRA mechanism, enabling targeted, scenario-specific adaptation with minimal overhead.
The key advantage of DriveVLA-M0 lies in the synergy between memory retrieval and test-time adaptation: memory provides access to past and external failure experiences, while TTT allows the VLA model to dynamically integrate this knowledge into its driving policy.
Together, they enable the model to handle scenarios that were previously failure-prone and improve robustness to distribution shifts without requiring large-scale retraining.

Extensive experiments on NAVSIMv1~\cite{navsimv1} and NAVSIMv2~\cite{navsimv2} demonstrate the effectiveness of our approach. DriveVLA-M0 achieves state-of-the-art performance across multiple metrics, reaching 94.1 PDMS on NAVSIMv1 and 47.0 EPDMS on NAVSIMv2, demonstrating superior results in safety-critical indicators such as collision avoidance and drivable area compliance. Moreover, we show that our framework scales naturally with external memory, enabling further gains through synthetic data augmentation without modifying the base model. Our contributions are as follows:
\begin{itemize}
\item We propose a failure-aware memory augmentation framework for VLA-based autonomous driving that explicitly targets poorly performing scenarios.
\item We design a structurally grounded retrieval mechanism that decouples static and dynamic scene features and integrate it with a decoupled LoRA-based test-time training strategy for efficient, targeted online adaptation.
\item We demonstrate that combining memory and TTT yields significant improvements in robustness, safety, and scalability across multiple benchmarks.
\end{itemize}


\begin{figure*}[t]
  \centering
  \includegraphics[width=0.85\textwidth]{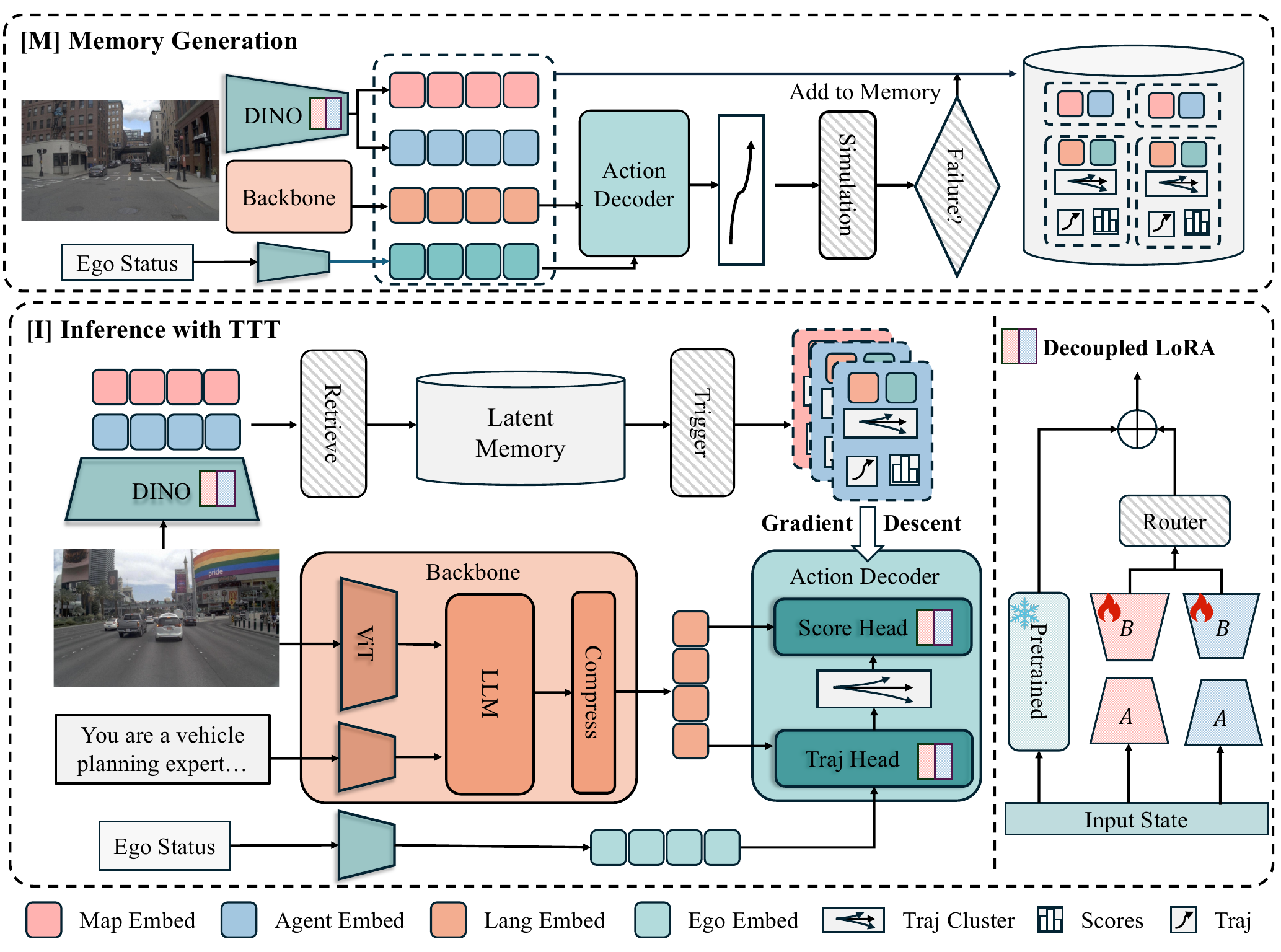}
  \caption{
    Overview of DriveVLA-M0. 
    (Top) [M] Memory Generation: Scenarios where the base model underperforms are identified via oracle simulation scoring, and their intermediate features are stored in latent memory $\mathbb{M}$. 
    (Bottom) [I] Inference with TTT: At test time, structurally similar cases are retrieved from $\mathbb{M}$ based on the current scene features, and the retrieved contents are used to fine-tune the Action Decoder via Decoupled LoRA. Specifically, cases retrieved by map features are utilized to adapt the Map (Static) LoRA branch, while cases retrieved by agent features are used to adapt the Agent (Dynamic) LoRA branch, enabling targeted and scenario-specific planning correction.
  }
  \label{fig:overview}
\end{figure*}

\section{Related Works}
 
\subsection{E2E Autonomous Driving and VLA Models}
 
E2E autonomous driving learns a unified policy mapping raw sensor inputs to planned trajectories~\cite{datascaling}. TransFuser~\cite{transfuser} introduces transformer-based sensor fusion for multi-modal feature integration, and UniAD~\cite{uniad} jointly optimizes perception, prediction, and planning through task-hierarchy design. DiffusionDrive~\cite{diffusiondrive} models multimodal action distributions with a truncated diffusion policy, and GoalFlow, Mimir, and MeanFuser~\cite{goalflow, mimir, meanfuser} reduce trajectory divergence by conditioning flow matching on a scene-selected goal point. TrajMoE~\cite{xing2025trajmoe} employs a mixture-of-experts planner with reinforcement-learning refinement to adapt trajectory priors across driving scenes. The proposal-and-score paradigm~\cite{vadv2,gtrs,ipad} like iPad generates diverse trajectory candidates and selects among them via multi-criteria scoring.
 
On the VLA side, DriveVLM~\cite{drivevlm} first applies large VLMs to provide semantic scene reasoning for driving. PlanAgent~\cite{planagent} studies MLLM planning. OmniDrive~\cite{omnidrive} grounds an LLM agent in 3D driving tasks with counterfactual reasoning annotations. ORION~\cite{orion} couples a memory-augmented LLM with trajectory generation and natural language explanations. OpenDriveVLA~\cite{opendrivevla} aligns 2D and 3D visual tokens in a unified semantic space for planning. 
VLMPlanner~\cite{vlmplanner} adopts a fast-slow system design to integrate VLM reasoning with efficient motion planning.
AutoVLA~\cite{autovla} integrates chain-of-thought reasoning and trajectory prediction in a single autoregressive model with GRPO-based reinforcement fine-tuning. AlphaDrive, DreamerAD, and ReCogDrive~\cite{alphadrive, dreamerad, recogdrive} combine reinforcement learning with VLM reasoning to handle safety-critical and long-tail scenarios. 
Recent works such as SERA, ELF-VLA, BeyondDrive, and SoAD~\cite{sera, elfvla, beyondimitation, soad} further exploit failure cases or safety values, requiring extensive post-training. In contrast, our method compresses failure contexts into a structured latent memory for direct test-time adaptation without offline retraining.
 
\subsection{Memory-Augmented Methods}
 
Memory augmentation enables models to store and reuse past experience at inference. In trajectory prediction, MANTRA~\cite{mantra} retrieves prototypical motion patterns from a memory bank, and MemoNet~\cite{memonet} captures retrospective intentions from historical agent trajectories for consistent multi-agent forecasting. In robotics, JARVIS-1~\cite{jarvis1} equips multimodal agents with episodic memory for open-world task execution, MemGen~\cite{memgen} accumulates latent memory without catastrophic forgetting, and STRAP~\cite{strap} retrieves sub-trajectories from a memory bank to augment policy learning. Titans~\cite{titans} proposes a neural memory module updated at inference, and Memento~\cite{memento} adapts LLM agents via external memory without modifying model weights. Agent memory studies management, retrieval, and utilization~\cite{d2skill, memchain, ucob}. In driving, EvoVLA~\cite{evovla} builds a self-evolving VLA through continuous experience accumulation,  MTRDrive~\cite{mtrdrive} stores prototypical corner-case scenarios for rapid retrieval, and MemoryVLA~\cite{memoryvla} introduces perceptual-cognitive memory into VLA planning. 
However, these methods primarily perform retrieval in the   
vision-language space, overlooking critical structural properties of driving scenes such as road topology and agent layouts, which can lead to retrieved cases that are semantically similar but structurally mismatched for safe decision-making.
 
\subsection{Test-Time Training and Parameter-Efficient Adaptation}
 
Test-time training (TTT) adapts model parameters on each test input to close the train-test distribution gap~\cite{sun2020ttt}. TENT~\cite{tent} minimizes prediction entropy on batch normalization statistics without labeled data, and TTT++~\cite{tttpp} combines contrastive self-supervision with feature alignment to handle natural distribution shifts. LoRA~\cite{lora} constrains weight updates to low-rank matrices for parameter-efficient fine-tuning. LoRA-TTT~\cite{lorattt} applies this to VLMs at test time, and Test-Time Low Rank Adaptation~\cite{testtimelora} extends it with confidence-based strategies for zero-shot generalization. STARAP~\cite{strap} shows that adapting with a small set of retrieved samples improves out-of-distribution performance in robotic policy learning. In autonomous driving, Centaur~\cite{centaur} performs test-time training by minimizing cluster entropy to reduce decision uncertainty. In contrast, we leverage memory to provide more explicit guidance signals. Meanwhile, our method enables post-training-free scaling through memory expansion.
\section{Method}

As illustrated in Figure~\ref{fig:overview}, inspired by the mechanism by which humans leverage failure-experience association for real-time error correction in complex decision-making~\cite{botvinick2001conflict}, we propose a Failure-Aware Latent Memory with Associative Correction architecture. Specially, we first train a \textbf{Base Model} via imitation learning, and a dedicated \textbf{Retrieve Model} using both static road information and dynamic surrounding-agent information. Then the deployment process consists of two core stages: \textbf{[M] Memory Generation} in the Offline stage and \textbf{[I] Inference with TTT} in the Online stage. During the Memory Generation stage, the base model is used to perform inference on past training or external data; failure scenarios are identified, and their intermediate road representations, surrounding-agent representations, and planning representations are written into the latent memory. During the Inference with TTT stage, the Retrieve Model retrieves structurally similar historical scenarios from memory, and a Decoupled LoRA architecture is employed to perform test-time training of the planning head, thereby enabling real-time correction in challenging driving situations.

Algorithm~\ref{alg:drivevla_m0} summarizes the complete DriveVLA-M0 pipeline and provides a procedural view of the two-stage design. The offline stage writes low-scoring scenarios into the latent memory, while the online stage retrieves structurally similar cases and activates Decoupled LoRA TTT only when the trigger condition is satisfied. The following subsections describe each component in detail.

\renewcommand{\algorithmiccomment}[1]{\hfill \texttt{//} #1}
\begin{algorithm}[t]
  \caption{\quad DriveVLA-M0: Failure-Aware Memory Augmentation with TTT}
  \label{alg:drivevla_m0}
  \scriptsize
  \begin{algorithmic}[1]
    \STATE \textbf{Input:} Base model $\mathcal{B}$, Retrieve Model $\mathcal{R}$, oracle scorer $\mathrm{PDM}(\cdot)$, memory $\mathbb{M}$, thresholds $\beta, \lambda$, TTT steps $S$, learning rate $\eta$
    \STATE \textbf{[M] Offline Memory Generation}
    \FOR{each scenario $s \in \mathcal{D}$}
      \STATE Run $\mathcal{B}$ to obtain trajectory $\hat{\tau}$ and intermediate features
      \STATE Extract retrieval keys $(F_{\mathrm{map}}, F_{\mathrm{agent}}) \leftarrow \mathcal{R}(I)$
      \STATE $\mathcal{Q}(\hat{\tau}) \leftarrow \mathrm{PDM}(\hat{\tau})$
      \IF{$\mathcal{Q}(\hat{\tau}) < \beta$}
        \STATE Store memory entry $(k, x, y)$ in $\mathbb{M}$ \COMMENT{keys, features, labels}
      \ENDIF
    \ENDFOR
    \STATE \textbf{[I] Online Inference with TTT}
    \STATE Encode current scene: $(F_{\mathrm{lang}}, F_{\mathrm{ego}}, \hat{\mathbb{T}}) \leftarrow \mathcal{B}_{\mathrm{enc}}(I,z)$
    \STATE Query memory using $(F^q_{\mathrm{map}}, F^q_{\mathrm{agent}}) \leftarrow \mathcal{R}(I)$
    \STATE Retrieve top-$k$ cases $\mathcal{C}_{\mathrm{map}}, \mathcal{C}_{\mathrm{agent}}$ with similarity above $\lambda$
    \IF{$\mathcal{C}_{\mathrm{map}} = \emptyset$ \textbf{and} $\mathcal{C}_{\mathrm{agent}} = \emptyset$}
      \RETURN base trajectory $\hat{\tau}^{0}$ \COMMENT{skip TTT}
    \ENDIF
    \STATE Initialize Map LoRA $\theta_{\mathrm{m}}$ and Agent LoRA $\theta_{\mathrm{a}}$
    \FOR{$i = 1$ to $S$}
      \STATE Optimize trajectory and score losses on retrieved cases
      \STATE $\theta \leftarrow \theta - \eta \nabla_{\theta}\mathcal{L}_{\mathrm{TTT}}(\mathcal{C}_{\mathrm{map}},\mathcal{C}_{\mathrm{agent}})$
    \ENDFOR
    \STATE Generate and score adapted trajectory candidates with the two LoRA branches
    \STATE Select $\hat{\tau}^{*}$ by the PDMS-style aggregated score
    \RETURN $\hat{\tau}^{*}$
  \end{algorithmic}
\end{algorithm}

\begin{figure*}[t]
  \centering
  \includegraphics[width=0.85\textwidth]{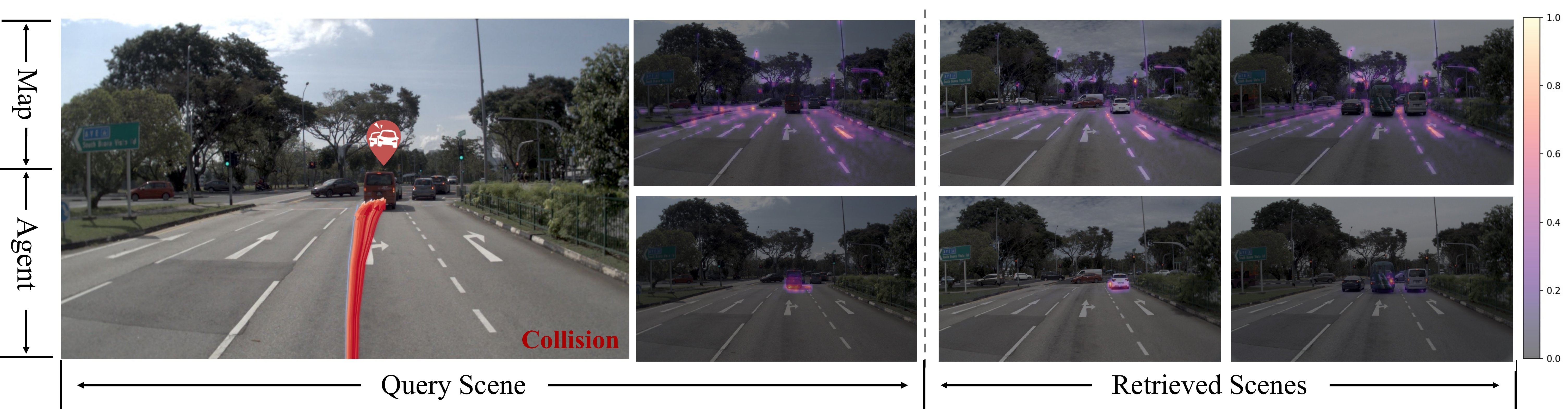}
    \caption{
    Attention maps of query and retrieved scenes. Top: map embedding; Bottom: agent embedding. Yellow indicates higher attention, and gray-purple indicates lower attention.
    }
  \label{fig:retrieve_demo}
\end{figure*}
\subsection{Base Model}

\noindent\textbf{VLM Backbone.}
Following RecogDrive~\cite{recogdrive}, the VLM backbone adopts InternVL3~\cite{internvl3} fine-tuned on large-scale autonomous-driving question-answering data. Given a front-view image $I \in \mathbb{R}^{C \times W \times H}$ and a system prompt $T$, the VLM encodes both modalities through separate encoders, maps them into a unified feature space, and feeds them into the subsequent LLM module. We extract the last-layer feature $h^{-1}$ of the LLM as the intermediate scene representation. However, the last-layer features $h^{-1}$ span $2800 \times 1536$ tokens, imposing prohibitive memory overhead. To mitigate this, we adopt a Q-Former-style~\cite{qformer} compression module with learnable queries $Q_{\text{cmp}} \in \mathbb{R}^{N \times D}$ ($N{=}16$, $D{=}256$) to condense the features via cross-attention:
\begin{equation}
  F_{\text{lang}} = \mathrm{Transformer}\!\left(Q_{\text{cmp}},\, \mathrm{Linear}(h^{-1}),\, \mathrm{Linear}(h^{-1})\right),
\end{equation}
This achieves a $1050~\times$ compression ratio, reducing the language-space representation $h^{-1}$ to a compact feature $F_{\text{lang}} \in \mathbb{R}^{N \times D}$.

\noindent\textbf{Action Decoder.}
Following score-based planning works~\cite{diffusiones,vadv2,gtrs,ipad}, our decoder operates in two stages. The Trajectory Head first compresses ego-status into $F_{\text{ego}} \in \mathbb{R}^{1 \times D}$ via a lightweight MLP, then decodes it jointly with $F_{\text{lang}}$ into a trajectory cluster $\hat{\mathbb{T}} \in \mathbb{R}^{M \times 8 \times 3}$, where $Q_{\text{ego}} \in \mathbb{R}^{M \times D}$ is a learnable embedding guiding $M$ diverse trajectory modes:
\begin{align}
  F_{\text{proposals}} &= \mathrm{Transformer}\!\left((Q_{\text{ego}} + F_{\text{ego}}),\; F_{\text{lang}},\; F_{\text{lang}}\right), \\
  \hat{\mathbb{T}} &= \mathrm{MLP}(F_{\text{proposals}}).
\end{align}
The Score Head subsequently re-encodes $\hat{\mathbb{T}}$ into $F'_{\text{proposals}}$, fuses it with $F_{\text{lang}}$, and produces $K$ sub-scores per proposal covering safety criteria such as collision rate and drivable area compliance:
\begin{align}
  F_{\text{scores}} &= \mathrm{Transformer}\!\left((F'_{\text{proposals}} + F_{\text{ego}}),\; F_{\text{lang}},\; F_{\text{lang}}\right), \\
  \hat{\mathbb{S}} &= \left\{\mathrm{MLP}_i(F_{\text{scores}}) \;\middle|\; i = 1,\ldots,K\right\},
\end{align}
where $\hat{\mathbb{S}} \in \mathbb{R}^{M \times K}$. The trajectory with the highest aggregated score is selected as the final output $\hat{\tau}$.
\subsection{Retrieve Model}
\label{sec: retrieve model}

How to retrieve similar scene is a key challenge in our memory system. Prior memory systems (e.g., MemoryVLA~\cite{memoryvla}) in embodied AI rely on vision–language features $F_{\text{lang}}$, which capture high-level semantics but may overlook structural properties such as road topology and agent layout when applied to autonomous driving scenarios. To capture structurally grounded scene representations, we train a Retrieve Model supervised on static road and dynamic agent information, producing decoupled map and agent embeddings as retrieval keys.



\begin{figure*}[th]
  \centering
  \includegraphics[width=0.8\textwidth]{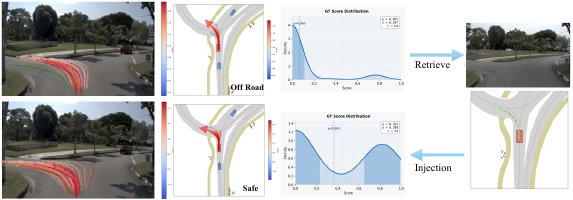}
  \caption{
    Visualization of trajectories distribution before (\textbf{top}) and after (\textbf{bottom}) TTT injection. Red trajectories indicate those the model is more likely to select, while green trajectories represent the human trajectory. The accompanying distribution plot shows the GT score density of the trajectory cluster, where a rightward shift corresponds to higher overall cluster quality.
  }
  \label{fig:injection}
\end{figure*}

\begin{figure}[!h]
  \centering
  \includegraphics[width=0.45\textwidth]{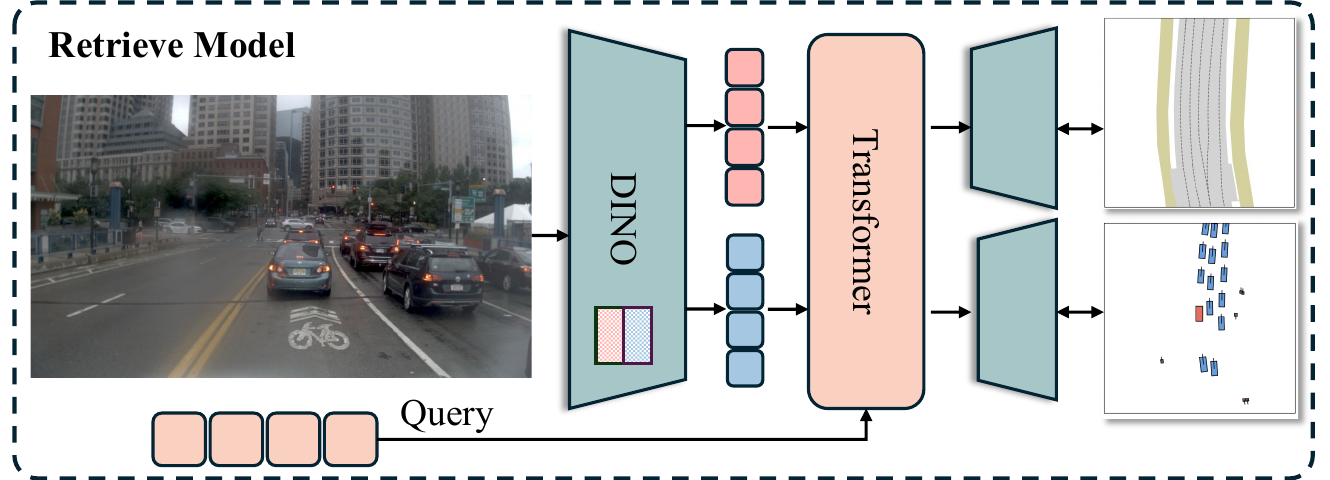}
  \caption{
    \textbf{Model Architecture of Retrieve Model.} 
  }
  \label{fig:retrieve_model}
\end{figure}

\noindent\textbf{Model Architecture.}
As illustrated in Figure~\ref{fig:retrieve_model}, we adopt a lightweight pre-trained DINOv2~\cite{dinov2} as the feature extractor and fine-tune it with LoRA~\cite{lora}. To enable independent adaptation of static road and dynamic agent representations in Sec.~\ref{sec: inference with TTT}, we employ a Decoupled LoRA strategy, separating the feature extraction into Map (static) and Agent (dynamic) branches.

Formally, the input $I$ is effectively routed through both LoRA branches in parallel, yielding $F_{\text{map}}$ and $F_{\text{agent}}$ simultaneously. The encoder produces disentangled features for map (static) and agent (dynamic) elements:
\begin{align}
F_{\text{map}}, F_{\text{agent}} = \mathrm{DINO}_{\text{LoRA}}(I),
\end{align}
where the Map branch captures static road structures and the Agent branch captures dynamic agent context.

Each feature map is then aggregated via a Transformer decoder with a learnable query, producing compact representations. Separate decoder heads further map these representations into occupancy grid outputs for map $\hat{\mathcal{M}}_{\text{map}}$ and agent $\hat{\mathcal{M}}_{\text{agent}}$.

\noindent\textbf{Training Recipe.}
Following the occupancy-grid supervision paradigm of Transfuser~\cite{transfuser}, we supervise the predicted maps $\hat{\mathcal{M}}_{\text{map}}$ and $\hat{\mathcal{M}}_{\text{agent}}$ using binary cross-entropy loss:
\begin{equation}
  \mathcal{L}_{\text{Retrieve}} = \mathrm{BCE}\!\left(\hat{\mathcal{M}}_{\text{map}},\, \mathcal{M}_{\text{map}}\right)
  + \alpha \cdot \mathrm{BCE}\!\left(\hat{\mathcal{M}}_{\text{agent}},\, \mathcal{M}_{\text{agent}}\right).
\end{equation}

We visualize the attention maps of map and agent embeddings in Figure~\ref{fig:retrieve_demo} and observe that map embeddings focus on road topology such as boundaries and lane markings, while agent embeddings primarily attend to front-facing vehicles. This decoupled design enables retrieval of structurally relevant failure cases conditioned on both scene geometry and agent context.
\subsection{Memory Generation}

The Memory Generation process writes case-level~\cite{memoryaiagents} episode from poorly performing scenarios which are drawn from training data or synthetic datasets into the latent memory. Each memory case encodes three categories of information: $k$: map $F_{\text{map}}$ and agent $F_{\text{agent}}$ embedding  for retrieval, $x$: intermediate representations of the Action Decoder for module input, and $y$: ground-truth labels for supervision.

\noindent\textbf{Latent Memory.}
Latent representations have been shown to effectively store multimodal information while maintaining a balance between storage efficiency and retrieval flexibility~\cite{memoryagent}. Building on this insight, we adopt a case-based latent memory design following approaches such as Momento~\cite{memento} and MemGen~\cite{memgen}. Each case in the latent memory $\mathbb{M}$ consists of:
(1) \emph{Retrieval keys}: static map feature $F_{\text{map}}$ and dynamic agent feature $F_{\text{agent}}$;
(2) \emph{Adaptation inputs}: language embedding $F_{\text{lang}}$, ego-status embedding $F_{\text{ego}}$, and the intermediate trajectory cluster $\hat{\mathbb{T}}$;
(3) \emph{Supervision targets}: the expert trajectory $\tau$ and the oracle-scored trajectory-cluster scores $\mathbb{S}$.

Formally, the memory is structured as:
\begin{equation}
  \mathbb{M} = \Bigl\{ \bigl(k_i,\; x_i,\; y_i\bigr) \Bigr\}_{i=1}^{N},
\end{equation}
where:
\begin{equation}
  k_i = \bigl(F^{(i)}_{\text{map}},\, F^{(i)}_{\text{agent}}\bigr), \quad
  x_i = \bigl(F^{(i)}_{\text{lang}},\, F^{(i)}_{\text{ego}},\, \hat{\mathbb{T}}^{(i)}\bigr), \quad
  y_i = \bigl(\tau^{(i)},\, \mathbb{S}^{(i)}\bigr).
\end{equation}

\noindent\textbf{Memory Construction.}
During memory construction, failure cases are identified from training or synthetic driving scenarios and stored as retrievable entries in the latent memory $\mathbb{M}$.
Given a scenario, two branches operate in parallel: the base model runs inference to produce the predicted trajectory $\hat{\tau}$ and caches intermediate representations $F_{\text{lang}}$, $F_{\text{ego}}$ and $\hat{\mathbb{T}}$, while the retrieval model $\mathrm{DINO}_{\text{LoRA}}$ extracts map and agent features $F_{\text{map}}$ and $F_{\text{agent}}$.
To assess planning quality, $\hat{\tau}$ is evaluated in simulation by the oracle scorer PDM~\cite{pdm}, which assigns sub-scores across $K$ safety and comfort criteria; these sub-scores are aggregated into $\mathcal{Q}(\hat{\tau}) \in [0,1]$ via a combination of multiplicative and additive operations.
A scenario is identified as a failure case when $\mathcal{Q}(\hat{\tau}) < \beta$, whereupon the associated features and supervision labels are written to $\mathbb{M}$:
\begin{equation}
    \mathbb{M} \leftarrow \mathbb{M} \cup \left\{ (F_{\text{lang}},\, F_{\text{ego}}, \, \hat{\mathbb{T}}),\; \left(F_{\text{map}},\, F_{\text{agent}}\right),\; (\tau,\; \mathbb{S}) \right\}
\end{equation}
where $\tau$ is the ground-truth expert trajectory serving as the Trajectory Head supervision label, and $\mathbb{S} = \{s_k\}_{k=1}^{K}$ serves as the Score Head supervision label.

In practice, to ensure scalability when incorporating large volumes of external data, we apply a cosine-similarity-based deduplication mechanism before writing each entry to check whether a sufficiently similar scenario already exists in $\mathbb{M}$, preventing redundancy and bounding the memory pool to a manageable size.

\begin{table*}[t]
\centering
\caption{\textbf{Comparison with state-of-the-art methods on Navtest (NAVSIMv1).} \textbf{Bold}: best. C: Camera. L: LiDAR.}
\label{tab:navsim_v1_main}
\begingroup
\fontsize{8.5pt}{9.4pt}\selectfont
\setlength{\tabcolsep}{5.0pt}
\renewcommand{\arraystretch}{0.9}
\begin{tabular}{l l l l | c c c c c | c}
\toprule
\textbf{Type} & \textbf{Method} & \textbf{Publication} & \textbf{Input} & \textbf{NC}$\uparrow$ & \textbf{DAC}$\uparrow$ & \textbf{EP}$\uparrow$ & \textbf{TTC}$\uparrow$ & \textbf{C}$\uparrow$ & \textbf{PDMS}$\uparrow$ \\
\midrule
\multirow{7}{*}{Classic E2E}
  & TransFuser~\cite{transfuser}         & TPAMI'23 & C (F+FL+FR) + L & 97.7 & 92.8 & 84.0 & 92.8 & 100 & 83.4 \\
  & World4Drive~\cite{world4drive}       & ICCV'25 & C (F+FL+FR)     & 97.4 & 94.3 & 79.9 & 92.8 & 100 & 85.1 \\
  & DiffusionDrive~\cite{diffusiondrive} & CVPR'25 & C (F+FL+FR) + L & 98.2 & 96.2 & 82.2 & 94.9 & 99.9 & 88.1 \\
  & Mimir~\cite{mimir}                   & RAL'25  & C (F+FL+FR)      & 98.2  & 97.5 & 83.6 & 94.6 & 100 & 89.3 \\
        & VADv2~\cite{vadv2}                   & ICLR'26 &C (Surround)    & 98.3 & 97.4 & 82.3 & 95.7 & 100 & 89.3 \\
  & GoalFlow~\cite{goalflow}             & CVPR'25 & C (F+FL+FR) + L & 98.4 & 98.3 & 85.0 & 94.6 & 100 & 90.3 \\
  & iPad~\cite{ipad}                     & arXiv'25 & C (Surround)     & 98.6 & 98.3 & 88.0 & 94.9 & 100 & 91.7   \\
  & Centaur~\cite{centaur}               & arXiv'25 & C (F+FL+FR) & 99.5 & 98.9 & 85.9 & 98.0 & 100 & 92.6   \\
  & DriveSuprim~\cite{yao2025drivesuprim} & AAAI'26 & C (F+FL+FR)     & 98.6 & 98.6 & 91.3 & 95.5 & 100 & 93.5   \\
\midrule
\multirow{5}{*}{VLA}
  & MTRDrive~\cite{mtrdrive}            & arXiv'25 & C (Front)    & 97.3 & 95.8 & 86.8 & 91.2 & 100 & 88.3 \\
  & AutoVLA~\cite{autovla}              & NeurIPS'25 & C (Surround) & 98.4 & 95.6 & 81.9 & 98.0 & 99.9 & 89.1 \\
  & DriveVLA-W0~\cite{drivevlaw0}       & ICLR'26 & C (Front)    &  98.7 & 99.1 & 83.3 & 95.3 & 99.3 & 90.2 \\
  & ReCogDrive~\cite{recogdrive}        & ICLR'26 & C (Front)    & 97.9 & 97.3 & 87.3 & 94.9 & 100 & 90.8 \\
  & ELF-VLA~\cite{elfvla}               & CVPR'26 & C (Front)             & 98.9 & 98.1 & 85.3 & 96.0 & 100 & 91.0 \\
  & MTDrive~\cite{mtdrive}              & arxiV'26 & C (Front)    & 97.5 & 98.2 & 90.6 & 91.8 & 99.8 & 91.1 \\
\midrule
\multirow{2}{*}{Ours}
  & DriveVLA-M0-Base                      & & C (Front)       & 99.0 & 97.7 & 89.5 & 95.0 & 99.9 & \textbf{92.3} \\
  & DriveVLA-M0-Scale                     & & C (Front)       & 99.1 & 98.1 & 90.2 & 98.5 & 99.9 & \textbf{94.1} \\
\bottomrule
\end{tabular}
\endgroup
\end{table*}

\subsection{Inference with TTT}
\label{sec: inference with TTT}


In this section, we describe how knowledge from $\mathbb{M}$ is injected into the base model to improve performance in challenging scenarios. Following prior works that use TTT for knowledge injection~\cite{strap,lorattt}, we retrieve analogous samples from the latent memory via the Retrieve Model, gated by a Trigger, and fine-tune the Action Decoder's planning head using the Decoupled LoRA mechanism which is same as Section~\ref{sec: retrieve model}. 

\noindent\textbf{Retrieve from Memory.}
At inference time, given the current front-view image $I$, we pass it through the Retrieve Model to extract the static road feature $F_{\text{map}}$ and the dynamic agent feature $F_{\text{agent}}$ as the retrieval key, which is used to query the latent memory $\mathbb{M}$. Retrieval is performed separately at the map level and the agent level, selecting the top-$k$ most similar cases for each. The retrieved features $x = (F_{\text{lang}}, F_{\text{ego}}, \hat{\mathbb{T}})$ and supervision labels $y = (\tau, \mathbb{S})$ are subsequently used for fine-tuning.

Since TTT adaptation is warranted only in scenarios with a high probability of planning failure, we introduce a Trigger mechanism based on cosine similarity:
\begin{equation}
  g = \begin{cases} 1, & \displaystyle\text{if } \frac{F^\top F^{*}}{\|F\|_2 \,\|F^{*}\|_2} > \lambda, \\ 0, & \text{otherwise,} \end{cases}
\end{equation}
where $g$ is the binary switch controlling whether TTT is activated, $F^* \in \{F^*_{\text{map}}, F^*_{\text{agent}}\}$ is the feature retrieved from $\mathbb{M}$.


\noindent\textbf{Inference with TTT.}
Once the relevant cases $\{x, y\}$ are retrieved, we fine-tune the $\mathrm{ActionDecoder}_{\text{LoRA}}$ via LoRA-based TTT, with LoRA weights re-initialized for each test scenario to ensure scenario-specific adaptation. The fine-tuning follows a decoupled design: map-retrieved cases are used to fine-tune the map (static) LoRA branch, while agent-retrieved cases fine-tune the agent (dynamic) LoRA branch, injecting map and agent knowledge separately into their respective modules.

At inference, both branches produce independent score predictions. For road-comprehension sub-scores such as drivable-area compliance, we adopt the Static LoRA predictions; for dynamic-capability sub-scores such as collision avoidance, we adopt the Dynamic LoRA predictions. This pathway-aware score fusion is applied across the trajectory cluster to select the optimal output $\hat{\tau}$.

In Figure~\ref{fig:injection}, we visualize the injection process. In the top panel, trajectories in the failure case are concentrated around zero, with almost no high-scoring options, preventing the model from selecting better trajectories. After retrieval and injection, the trajectory cluster quality improves and the GT score distribution shifts right, allowing model select safe trajectory from this cluster.
\section{Experiments}

\begin{table*}[t]
\centering
\setlength{\tabcolsep}{6pt}
\renewcommand{\arraystretch}{0.95}
\caption{\textbf{Comparision with state-of-the-art methods on the Navhard (NAVSIMv2).} All methods rely on sensor data for planning.}
\vspace{0.1in}
\scriptsize
\resizebox{0.75\textwidth}{!}{
\begin{tabular}{c | c | c c c c c c c c c | c}
\toprule
Method
& Stage
& NC
& DAC
& DDC
& TLC
& EP
& TTC
& LK
& HC
& EC
& EPDMS \\
\midrule

LTF~\citep{transfuser} &
\makecell{Stage 1 \\ Stage 2} &
\makecell{96.2 \\ 77.7} & \makecell{79.5 \\ 70.2} & \makecell{99.1 \\ 84.2} & \makecell{99.5 \\ 98.0} &
\makecell{84.1 \\ 85.1} & \makecell{95.1 \\ 75.6} & \makecell{94.2 \\ 45.4} & \makecell{97.5 \\ 95.7} &
\makecell{79.1 \\ 75.9} & 23.1 \\
\midrule

DiffusionDrive~\citep{diffusiondrive} &
\makecell{Stage 1 \\ Stage 2} &
\makecell{96.8 \\ 80.3} & \makecell{88.2 \\ 74.4} & \makecell{99.3 \\ 86.1} & \makecell{99.3 \\ 98.4} &
\makecell{84.5 \\ 87.4} & \makecell{94.6 \\ 76.9} & \makecell{95.5 \\ 50.4} & \makecell{97.5 \\ 95.5} &
\makecell{79.1 \\ 69.4} & 28.9 \\
\midrule

GoalFlow~\citep{goalflow} &
\makecell{Stage 1 \\ Stage 2} &
\makecell{96.0 \\ 79.4} & \makecell{92.6 \\ 78.2} & \makecell{99.3 \\ 86.4} & \makecell{99.3 \\ 97.7} &
\makecell{84.0 \\ 86.5} & \makecell{95.7 \\ 76.0} & \makecell{97.1 \\ 45.5} & \makecell{97.5 \\ 94.4} &
\makecell{40.4 \\ 40.4} & 28.7 \\
\midrule

Mimir~\citep{mimir} &
\makecell{Stage 1 \\ Stage 2} &
\makecell{95.6 \\ 80.6} & \makecell{92.2 \\ 77.1} & \makecell{99.7 \\ 89.3} & \makecell{99.5 \\ 97.7} &
\makecell{84.0 \\ 86.4} & \makecell{94.6 \\ 77.3} & \makecell{98.0 \\ 48.8} & \makecell{97.5 \\ 94.5} &
\makecell{78.2 \\ 64.0} & 34.6 \\
\midrule

DriveSuprim~\citep{yao2025drivesuprim} &
\makecell{Stage 1 \\ Stage 2} &
\makecell{98.7 \\ 89.5} & \makecell{98.0 \\ 89.6} & \makecell{99.1 \\ 92.9} & \makecell{99.8 \\ 98.5} &
\makecell{75.9 \\ 78.9} & \makecell{98.7 \\ 86.4} & \makecell{94.7 \\ 55.3} & \makecell{97.6 \\ 96.5} &
\makecell{49.8 \\ 52.7} & 44.7 \\
\midrule

GTRS-Dense~\citep{gtrs} &
\makecell{Stage 1 \\ Stage 2} &
\makecell{98.9 \\ 91.5} & \makecell{98.2 \\ 90.8} & \makecell{99.8 \\ 94.7} & \makecell{99.6 \\ 98.5} &
\makecell{73.9 \\ 70.8} & \makecell{98.9 \\ 90.1} & \makecell{95.3 \\ 55.4} & \makecell{97.3 \\ 97.2} &
\makecell{40.0 \\ 54.2} & 45.3 \\
\midrule

DriveVLA-M0-Base &
\makecell{Stage 1 \\ Stage 2} &
\makecell{98.9 \\ 91.1} & \makecell{96.2 \\ 89.5} & \makecell{99.7 \\ 93.1} & \makecell{100.0 \\ 98.6} &
\makecell{73.8 \\ 65.5} & \makecell{99.1 \\ 89.2} & \makecell{94.2 \\ 52.7} & \makecell{97.3 \\ 98.6} &
\makecell{64.4 \\ 72.1} & \textbf{47.0} \\
\bottomrule
\end{tabular}
}
\label{table:navsim_v2}
\end{table*}

\begin{table*}[t]
\centering
\setlength{\tabcolsep}{4.2pt}
\renewcommand{\arraystretch}{0.95}
\caption{\textbf{Comparison with representative methods on Navtest using the NAVSIMv2 EPDMS metrics.} All results follow the non-reactive evaluation protocol. \textbf{Bold}: best EPDMS.}
\vspace{0.1in}
\scriptsize
\resizebox{0.88\textwidth}{!}{
\begin{tabular}{l | c c c c c c c c c | c}
\toprule
Method
& NC
& DAC
& DDC
& TLC
& EP
& TTC
& LK
& HC
& EC
& EPDMS \\
\midrule
TransFuser~\citep{transfuser}
& 96.9 & 89.9 & 97.8 & 99.7 & 87.1 & 95.4 & 92.7 & 98.3 & 87.2 & 76.7 \\
DiffusionDrive~\citep{diffusiondrive}
& 98.2 & 95.9 & 99.4 & 99.8 & 87.5 & 97.3 & 96.8 & 98.3 & 87.7 & 84.5 \\
GoalFlow~\citep{goalflow}
& 98.4 & 98.2 & 99.5 & 99.8 & 88.1 & 97.4 & 96.7 & 98.2 & 57.9 & 87.3 \\
DriveVLA-W0~\citep{drivevlaw0}
& 98.5 & 99.1 & 98.0 & 99.7 & 86.4 & 98.1 & 93.2 & 97.9 & 58.9 & 86.1 \\
ReCogDrive~\citep{recogdrive}
& 98.3 & 95.2 & 99.5 & 99.8 & 87.1 & 97.5 & 96.6 & 98.3 & 86.5 & 83.6 \\
\midrule
DriveVLA-M0
& 98.0 & 97.9 & 99.6 & 99.5 & 89.7 & 97.9 & 96.3 & 98.2 & 75.5 & \textbf{88.9} \\
\bottomrule
\end{tabular}
}
\label{tab:navsim_v2_navtest}
\end{table*}
 
\subsection{Benchmarks}
We evaluate DriveVLA-M0 on two complementary benchmarks spanning non-reactive open-loop simulation and reactive pseudo closed-loop evaluation.
 
\noindent\textbf{NAVSIMv1}~\cite{navsimv1}.
NAVSIMv1 is a non-reactive, data-driven autonomous driving benchmark built on sensor data from nuPlan~\cite{nuplan} and OpenScene~\cite{openscene}. It computes simulation-based metrics by unrolling bird’s-eye-view (BEV) abstractions of scenes over a short simulation horizon, during which the evaluated policy and the environment do not interact. This decoupling enables open-loop metric computation while achieving better correlation with closed-loop evaluations than traditional displacement errors. The primary metric is the PDMS, which comprises five sub-scores. No-at-fault Collision (NC) penalizes collisions caused by the ego vehicle. Drivable Area Compliance (DAC) measures whether the ego vehicle remains within the drivable region. Time-to-Collision (TTC) assesses the safety margin relative to surrounding agents. Comfort (C) penalizes excessive jerk or acceleration. Ego Progress (EP) rewards forward progress along the planned route.
 
\noindent\textbf{NAVSIMv2}~\cite{navsimv2}.
NAVSIM v2 extends v1 with a richer metric and a pseudo closed-loop evaluation protocol. The metric is the Extended PDM Score (EPDMS), which adds four sub-scores to PDMS. Lane Keeping (LK) checks whether the ego vehicle stays within its current lane. Driving Direction Compliance (DDC) penalizes driving against the intended traffic direction. Traffic Light Compliance (TLC) enforces adherence to traffic signal states. Extended Comfort (EC) imposes stricter kinematic smoothness constraints. Evaluation follows a two-stage pipeline: Stage~1 scores the initial planned trajectory, and Stage~2 scores the planner on pre-computed follow-up scenes that branch from the Stage~1 outcome, better approximating long-horizon closed-loop behavior.
 

\subsection{Main Results}
 
\textbf{NAVSIMv1.}
Table~\ref{tab:navsim_v1_main} compares DriveVLA-M0 with state-of-the-art methods on the NAVSIMv1. We categorize prior approaches into two groups: classic E2E planners and VLA-based planners. With both map and agent retrieval, DriveVLA-M0-Base achieves a PDMS score of 92.3, outperforming prior VLA-based methods and remaining competitive with the strongest existing E2E approaches, including Centaur, which also adopts a TTT paradigm.

\noindent\textbf{NAVSIMv2.}
As shown in Table~\ref{tab:navsim_v2_navtest}, DriveVLA-M0 achieves the best EPDMS of 88.9 on Navtest, exceeding GoalFlow by 1.6 points. It also obtains the highest EP and DDC scores, indicating improved driving progress while maintaining strong directional compliance. We further evaluate our method on the more challenging Navhard benchmark, as shown in Table~\ref{table:navsim_v2}. Since Navhard is newly introduced, many VLA-based methods have not yet reported results on it; we therefore primarily compare against strong end-to-end baselines. DriveVLA-M0 improves key metrics such as NC and DAC over direct trajectory generation methods including Transfuser~\cite{transfuser}, while outperforming scoring-based methods such as GTRS~\cite{gtrs} in trajectory continuity EC. Overall, our approach achieves more balanced performance across metrics than both paradigms.


\textbf{Scaling with Memory Size.}
To validate scalability, we use SimScale~\cite{simscale} to synthesize additional scenarios and expand the latent memory to 10K cases without retraining, as shown in Table~\ref{tab:navsim_v1_main}. Since new failure cases can be directly added to memory and retrieved at inference time, the enlarged memory provides more relevant experiences for TTT-based correction, leading to consistent performance gains.
 
\subsection{Ablation Studies}
 
We conduct ablation studies on NAVSIMv1 to validate the effectiveness of each component. Unless otherwise specified, all ablation experiments are based on the same base model (PDMS $= 91.0$).

\textbf{Effect of Retrieval Strategy.}
Table~\ref{tab:ablation_search} ablates the retrieval key design. Using language embedding $F_{\text{lang}}$ for retrieval obtains 90.7 PDMS, slightly below the 91.0 PDMS base model, supporting our hypothesis that language-space representations are less effective at capturing the intrinsic characteristics of driving scenes for reliable retrieval. Switching to map-only retrieval using $F_{map}$ achieves 91.7 PDMS, and map+agent embedding $F_{agent}$ retrieval reaches 92.3 PDMS, improving over the base model by \textbf{1.3 PDMS}. Compared with map-only retrieval, adding agent embeddings improves NC, EP, and TTC by \textbf{0.5}, \textbf{0.5}, and \textbf{0.5}, respectively. This validates our design of decoupling map and agent embedding: the map embedding enables the planning head to focus on road topology for DAC and EP improvement, while the agent embedding enables focused optimization of collision-related metrics like NC and TTC.

\begin{table}[t]
\centering
\caption{Ablation on retrieval strategy. $\dagger$ denotes evaluation without memory.}
\small 
\label{tab:ablation_search}
\begin{tabular}{lccccc|c}
\toprule
Search Type & NC & DAC & EP & TTC & C & PDMS \\
\midrule
Base Model$^\dagger$       & 98.4 & 97.1 & 87.7 & 95.2 & 97.6 & 91.0 \\
\midrule
Lang         & 98.0 & 97.3 & 88.1 & 93.9 & 100.0 & 90.7 \\
Map          & 98.4 & 97.7 & 89.1 & 94.5 & 99.9 & 91.7 \\
Map + Agent  & 98.9 & 97.7 & 89.6 & 95.0 & 99.9 & \textbf{92.3} \\
\bottomrule
\end{tabular}
\end{table}

 
\textbf{Effect of Knowledge Injection Strategy.} Table~\ref{tab:ablation_injection} compares different approaches for injecting retrieved memory into the model, grouped into offline and TTT paradigms. In the offline setting, we perform post-training on the base model using its failure cases. Specifically, we add LoRA only to the Action Decoder and continue training on the error samples for 10 epochs. This slightly improves over the 91.0 PDMS base model, achieving a PDMS of 91.2. Although leveraging failure cases improves performance, it still falls short of TTT-based injection. We attribute this to a distribution mismatch, as the model is trained on a fixed data mixture without scenario-specific adaptation. These results indicate that offline methods cannot fully exploit the corrective potential of memory.
In contrast, TTT enables scenario-specific adaptation by fine-tuning the model on retrieved cases at inference time. Full Action Decoder TTT achieves the best performance with 92.4 PDMS, while Decoupled LoRA TTT reaches a comparable 92.3 PDMS. This efficiency–performance trade-off demonstrates that low-rank adaptation at test time can match full fine-tuning while significantly reducing computational overhead.



\begin{table}[t]
\centering
\caption{Ablation on knowledge injection strategy. All non-base methods use map + agent retrieval. $\dagger$ denotes evaluation without memory.}
\label{tab:ablation_injection}

\begin{tabular}{lcccccc}
\toprule
Injection Type & NC & DAC & EP & TTC & C & PDMS \\
\midrule
Base Model$^\dagger$       & 98.4 & 97.1 & 87.7 & 95.2 & 97.6 & 91.0 \\
\midrule
Offline (10 ep.)    & 98.1 & 97.5 & 88.6 & 94.2 & 99.9 & 91.2 \\
TTT Full         & 99.0 & 97.8 & 89.6 & 95.1 & 99.9 & 92.4 \\
TTT LoRA       & 99.0 & 97.8 & 89.5 & 95.0 & 99.9 & \textbf{92.3} \\
\bottomrule
\end{tabular}

\end{table}

\textbf{Effect of Trigger Threshold.}
In Section~\ref{sec: inference with TTT}, we use a trigger to select scenarios whose cosine similarity exceeds a threshold $\lambda$ and retrieve the top-$k$ cases from the latent memory for TTT. The Trigger thus controls the fraction of scenarios that undergo TTT, balancing correction coverage against the risk of introducing noise from irrelevant retrievals. Table~\ref{tab:ablation_trigger} reports results under different cosine similarity thresholds $\lambda$ using map-only retrieval.

When $\lambda = 0.7$, the threshold is overly permissive: many dissimilar scenarios are selected, leading to noisy supervision that degrades NC and TTC. In contrast, when $\lambda = 0.99$, the threshold becomes too restrictive: very few scenarios trigger adaptation, and performance drops to 89.4 PDMS, below the 91.0 PDMS base model. The best performance is achieved at $\lambda = 0.9$, reaching 91.7 PDMS. These results validate our design principle that selective adaptation based on high-confidence structural similarity is more effective than indiscriminate test-time fine-tuning.


\begin{table}[t]
\centering
\caption{Ablation on Trigger threshold $\lambda$ (map-only retrieval).}
\label{tab:ablation_trigger}
\begin{tabular}{lccccc|c}
\toprule
$\lambda$ & NC & DAC & EP & TTC & C & PDMS \\
\midrule
0.70 & 98.0 & 96.9 & 88.1 & 93.8 & 99.9 & 90.4 \\
0.90 & 98.4 & 97.7 & 89.1 & 94.5 & 99.9 & \textbf{91.7} \\
0.95 & 98.1 & 97.6 & 88.8 & 94.3 & 99.9 & 91.4 \\
0.99 & 98.1 & 96.9 & 85.2 & 94.3 & 99.9 & 89.4 \\
\bottomrule
\end{tabular}
\end{table}

\begin{table}[h]
\centering
\caption{Latency breakdown of DriveVLA-M0 components (ms).}
\label{tab:latency}
\setlength{\tabcolsep}{3.5pt} 
\begin{tabular}{lcccc}
\toprule
Component & Retrieve & Forward & Backward LoRA & Backward Full \\
\midrule
Time (ms) & 15.19 & 30.79 & 26.44 & 55.42 \\
\bottomrule
\end{tabular}
\end{table}

\textbf{Efficiency Analysis.}
Table~\ref{tab:latency} summarizes the inference-time efficiency of DriveVLA-M0. All measurements are conducted on a single NVIDIA H20 GPU with a latent memory containing 4,000 cases.
The retrieval stage is highly efficient: querying the memory takes only 15.19 ms. A single forward pass requires 30.79 ms. For TTT, the backward pass with Decoupled LoRA takes 26.44 ms, compared to 55.42 ms for full Action Decoder, demonstrating a significant reduction in latency cost.
Overall, these results show that DriveVLA-M0 introduces only modest overhead during inference, while enabling effective injection when TTT is triggered.

\section{Conclusion}
We present DriveVLA-M0, a failure-aware memory augmentation framework that enhances VLA-based autonomous driving with structured retrieval and test-time adaptation. Unlike prior methods that rely on static model parameters and language-level embeddings, DriveVLA-M0 explicitly targets failure-prone scenarios and retrieves structurally relevant cases based on road topology and agent layout. A Decoupled LoRA TTT mechanism injects the retrieved knowledge into the Action Decoder in a scenario-specific manner, activating only when necessary to minimize overhead. Experiments on NAVSIMv1 and NAVSIMv2 demonstrate consistent improvements over prior methods, with further gains achievable by expanding the memory pool with synthetic data without retraining the base model.

\bibliographystyle{ACM-Reference-Format}
\bibliography{drivemem_references}

@article{drivevlm,
  author  = {Xiaoyu Tian and Junru Gu and Bailin Li and Yicheng Liu and Yang Wang and Zhiyong Zhao and Kun Zhan and Peng Jia and Xianpeng Lang and Hang Zhao},
  title   = {DriveVLM: The Convergence of Autonomous Driving and Large Vision-Language Models},
  journal = {CVPR},
  year    = {2024}
}

@article{world4drive,
  author  = {Yupeng Zheng and Pengxuan Yang and Zebin Xing and Qichao Zhang and Yuhang Zheng and Yinfeng Gao and Pengfei Li and Teng Zhang and Zhongpu Xia and Peng Jia and Dongbin Zhao},
  title   = {World4Drive: End-to-End Autonomous Driving via Intention-aware Physical Latent World Model},
  journal = {ICCV},
  year    = {2025}
}

@misc{mtdrive,
      title={MTDrive: Multi-turn Interactive Reinforcement Learning for Autonomous Driving}, 
      author={Xidong Li and Mingyu Guo and Chenchao Xu and Bailin Li and Wenjing Zhu and Yangang Zou and Rui Chen and Zehuan Wang},
      year={2026},
      journal={arXiv:2601.22930}
}

@article{simscale,
  author  = {Tian, Haochen and Li, Tianyu and Liu, Haochen and Yang, Jiazhi and Qiu, Yihang and Li, Guang and Wang, Junli and Gao, Yinfeng and Zhang, Zhang and Wang, Liang and Ye, Hangjun and Tan, Tieniu and Chen, Long and Li, Hongyang},
  title   = {SimScale: Learning to Drive via Real-World Simulation at Scale},
  journal = {CVPR},
  year    = {2026}
}

@article{vlmplanner,
  author  = {Zhipeng Tang and Sha Zhang and Jiajun Deng and Chenjie Wang and Guoliang You and Yuting Huang and Xinrui Lin and Yanyong Zhang},
  title   = {VLMPlanner: Integrating Visual Language Models with Motion Planning},
  journal = {ACM MM},
  year    = {2025}
}

@article{sera,
  author  = {Xinyu Xia and Xingjun Ma and Yunfeng Hu and Ting Qu and Hong Chen and Xun Gong},
  title   = {From Failures to Fixes: LLM-Driven Scenario Repair for Self-Evolving Autonomous Driving},
  journal = {ACM MM},
  year    = {2025}
}

@article{orion,
  author  = {Haoyu Fu and Diankun Zhang and Zongchuang Zhao and Jianfeng Cui and Dingkang Liang and Chong Zhang and Dingyuan Zhang and Hongwei Xie and Bing Wang and Xiang Bai},
  title   = {ORION: A Holistic End-to-End Autonomous Driving Framework by Vision-Language Instructed Action Generation},
  journal = {ICCV},
  year    = {2025}
}

@article{omnidrive,
  author  = {Shihao Wang and Zhiding Yu and Xiaohui Jiang and Shiyi Lan and Min Shi and Nadine Chang and Jan Kautz and Ying Li and Jose M. Alvarez},
  title   = {OmniDrive: A Holistic Vision-Language Dataset for Autonomous Driving with Counterfactual Reasoning},
  journal = {CVPR},
  year    = {2025}
}

@article{autovla,
  author  = {Zewei Zhou and Tianhui Cai and Seth Z. Zhao and Yun Zhang and Zhiyu Huang and Bolei Zhou and Jiaqi Ma},
  title   = {AutoVLA: A Vision-Language-Action Model for End-to-End Autonomous Driving with Adaptive Reasoning and Reinforcement Fine-Tuning},
  journal = {NeurIPS},
  year    = {2025}
}

@article{memoryvla,
  author  = {Hao Shi and Bin Xie and Yingfei Liu and Lin Sun and Fengrong Liu and Tiancai Wang and Erjin Zhou and Haoqiang Fan and Xiangyu Zhang and Gao Huang},
  title   = {MemoryVLA: Perceptual-Cognitive Memory in Vision-Language-Action Models for Robotic Manipulation},
  journal = {ICLR},
  year    = {2026}
}

@article{mimir,
  author  = {Zebin Xing and Yupeng Zheng and Qichao Zhang and Zhixing Ding and Pengxuan Yang and Songen Gu and Zhongpu Xia and Dongbin Zhao},
  title   = {Mimir: Hierarchical Goal-Driven Diffusion With Uncertainty Propagation for End-to-End Autonomous Driving},
  journal = {IEEE Robotics and Automation Letters},
  volume  = {11},
  number  = {2},
  pages   = {2178--2185},
  year    = {2026},
  doi     = {10.1109/LRA.2025.3641129}
}

@article{evovla,
  author = {Zeting Liu and Zida Yang and Zeyu Zhang and Hao Tang},
  title = {EvoVLA: Self-Evolving Vision-Language-Action Model},
  journal = {arXiv:2511.16166},
  year = {2025}
}

@article{drivevlaw0,
  author  = {Yingyan Li and Shuyao Shang and Weisong Liu and Bing Zhan and Haochen Wang and Yuqi Wang and Yuntao Chen and Xiaoman Wang and Yasong An and Chufeng Tang and Lu Hou and Lue Fan and Zhaoxiang Zhang},
  title   = {DriveVLA-W0: World Models Amplify Data Scaling Law in Autonomous Driving},
  journal = {ICLR},
  year    = {2026}
}

@article{mtrdrive,
  author = {Yiming Luo and Haoran Zhang and Xiaosong Chen and Feng Zhao and Zhiqi Wang},
  title = {MTRDrive: Memory-Tool Synergistic Reasoning for Robust Autonomous Driving in Corner Cases},
  journal = {arXiv:2502.14329},
  year = {2025}
}

@article{alphadrive,
  author = {Bo Jiang and Shaoyu Chen and Qian Zhang and Wenyu Liu and Xinggang Wang},
  title = {AlphaDrive: Unleashing the Power of VLMs in Autonomous Driving via Reinforcement Learning and Reasoning},
  journal = {arXiv:2503.07608},
  year = {2025}
}

@article{recogdrive,
  author  = {Yongkang Li and Kaixin Xiong and Xiangyu Guo and Fang Li and Sixu Yan and Gangwei Xu and Lijun Zhou and Long Chen and Haiyang Sun and Bing Wang and Kun Ma and Guang Chen and Hangjun Ye and Wenyu Liu and Xinggang Wang},
  title   = {ReCogDrive: A Reinforced Cognitive Framework for End-to-End Autonomous Driving},
  journal = {ICLR},
  year    = {2026}
}

@article{opendrivevla,
  author = {Xingcheng Zhou and Xuyuan Han and Feng Yang and Yunpu Ma and Volker Tresp and Alois Knoll},
  title = {OpenDriveVLA: Towards End-to-end Autonomous Driving with Large Vision-Language-Action Model},
  journal = {arXiv:2503.23463},
  year = {2025}
}

@article{echovla,
  author = {Min Lin and Xiwen Liang and Bingqian Lin and Liu Jingzhi and Zijian Jiao and Kehan Li and Yu Sun and Weijia Liufu and Yuhan Ma and Yuecheng Liu and Shen Zhao and Yuzheng Zhuang and Xiaodan Liang},
  title = {EchoVLA: Robotic Vision-Language-Action Model with Synergistic Declarative Memory for Mobile Manipulation},
  journal = {arXiv:2511.18112},
  year = {2025}
}

@article{elfvla,
  author  = {Yuechen Luo and Qimao Chen and Fang Li and Shaoqing Xu and Jaxin Liu and Ziying Song and Zhi-xin Yang and Fuxi Wen},
  title   = {Unleashing VLA Potentials in Autonomous Driving via Explicit Learning from Failures},
  journal = {CVPR},
  year    = {2026}
}

@article{uniad,
  author  = {Yihan Hu and Jiazhi Yang and Li Chen and Keyu Li and Chonghao Sima and Xizhou Zhu and Siqi Chai and Senyao Du and Tianwei Lin and Wenhai Wang and Lewei Lu and Xiaosong Jia and Qiang Liu and Jifeng Dai and Yu Qiao and Hongyang Li},
  title   = {Planning-Oriented Autonomous Driving},
  journal = {CVPR},
  year    = {2023}
}

@article{vadv2,
  author  = {Shaoyu Chen and Bo Jiang and Hao Gao and Bencheng Liao and Qing Xu and Qian Zhang and Chang Huang and Wenyu Liu and Xinggang Wang},
  title   = {VADv2: End-to-End Vectorized Autonomous Driving via Probabilistic Planning},
  journal = {ICLR},
  year    = {2026}
}

@article{diffusiondrive,
  author  = {Bencheng Liao and Shaoyu Chen and Haoran Yin and Bo Jiang and Cheng Wang and Sixu Yan and Xinbang Zhang and Xiangyu Li and Ying Zhang and Qian Zhang and Xinggang Wang},
  title   = {DiffusionDrive: Truncated Diffusion Model for End-to-End Autonomous Driving},
  journal = {CVPR},
  year    = {2025}
}

@article{goalflow,
  author  = {Zebin Xing and Xingyu Zhang and Yang Hu and Bo Jiang and Tong He and Qian Zhang and Xiaoxiao Long and Wei Yin},
  title   = {GoalFlow: Goal-Driven Flow Matching for Multimodal Trajectories Generation in End-to-End Autonomous Driving},
  journal = {CVPR},
  year    = {2025}
}

@article{ipad,
  author = {Ke Guo and Haochen Liu and Xiaojun Wu and Jia Pan and Chen Lv},
  title = {iPad: Iterative Proposal-centric End-to-End Autonomous Driving},
  journal = {arXiv:2505.15111},
  year = {2025}
}

@article{centaur,
  author = {Chonghao Sima and Kashyap Chitta and Zhiding Yu and Shiyi Lan and Ping Luo and Andreas Geiger and Hongyang Li and Jose M. Alvarez},
  title = {Centaur: Robust End-to-End Autonomous Driving with Test-Time Training},
  journal = {arXiv:2503.11650},
  year = {2025}
}

@article{mantra,
  author  = {Francesco Marchetti and Federico Becattini and Lorenzo Seidenari and Alberto Del Bimbo},
  title   = {MANTRA: Memory Augmented Networks for Multiple Trajectory Prediction},
  journal = {CVPR},
  year    = {2020}
}

@article{memonet,
  author  = {Chenxin Xu and Weibo Mao and Wenjun Zhang and Siheng Chen},
  title   = {Remember Intentions: Retrospective-Memory-based Trajectory Prediction},
  journal = {CVPR},
  year    = {2022}
}

@article{jarvis1,
  author  = {Zihao Wang and Shaofei Cai and Anji Liu and Yonggang Jin and Jinbing Hou and Bowei Zhang and Haowei Lin and Zhaofeng He and Zilong Zheng and Yaodong Yang and Xiaojian Ma and Yitao Liang},
  title   = {JARVIS-1: Open-World Multi-task Agents with Memory-Augmented Multimodal Language Models},
  journal = {NeurIPS},
  year    = {2023}
}

@article{memgen,
  author  = {Guibin Zhang and Muxin Fu and Shuicheng Yan},
  title   = {MemGen: Weaving Generative Latent Memory for Self-Evolving Agents},
  journal = {ICLR},
  year    = {2026}
}

@article{memento,
  author = {Huichi Zhou and Yihang Chen and Siyuan Guo and Xue Yan and Kin Hei Lee and Zihan Wang and Ka Yiu Lee and Guchun Zhang and Kun Shao and Linyi Yang and Jun Wang},
  title = {Memento: Fine-tuning LLM Agents without Fine-tuning LLMs},
  journal = {arXiv:2508.16153},
  year = {2025}
}

@article{titans,
  author  = {Ali Behrouz and Peilin Zhong and Vahab Mirrokni},
  title   = {Titans: Learning to Memorize at Test Time},
  journal = {NeurIPS},
  year    = {2024}
}

@article{strap,
  author  = {Martin Memmel and Zhiqi Wang and Xiaosong Chen and others},
  title   = {STRAP: Robot Sub-Trajectory Retrieval for Augmented Policy Learning},
  journal = {ICLR},
  year    = {2025}
}

@article{sun2020ttt,
  author  = {Yu Sun and Xiaolong Wang and Zhuang Liu and John Miller and Alexei A. Efros and Moritz Hardt},
  title   = {Test-Time Training with Self-Supervision for Generalization under Distribution Shifts},
  journal = {ICML},
  year    = {2020}
}

@article{tent,
  author  = {Dequan Wang and Evan Shelhamer and Shaoteng Liu and Bruno Olshausen and Trevor Darrell},
  title   = {TENT: Fully Test-Time Adaptation by Entropy Minimization},
  journal = {ICLR},
  year    = {2021}
}

@article{tttpp,
  author  = {Yuejiang Liu and Supreeth Kothandaraman and Kiana Wadhwa and others},
  title   = {TTT++: When Does Self-Supervised Test-Time Training Fail or Thrive?},
  journal = {NeurIPS},
  year    = {2021}
}

@article{lorattt,
  author  = {Yuto Kojima and Jiarui Xu and Xueyan Zou and Xiaolong Wang},
  title   = {LoRA-TTT: Low-Rank Test-Time Training for Vision-Language Models},
  journal = {ICML},
  year    = {2025}
}

@article{testtimelora,
  author  = {Syed Talal Imam and Marwan Afifi and Joydeep Ghosh and others},
  title   = {Test-Time Low Rank Adaptation via Confidence Maximization for Zero-Shot Generalization of Vision-Language Models},
  journal = {WACV},
  year    = {2025}
}

@article{navsimv1,
  author  = {Daniel Dauner and Marcel Hallgarten and Tianyu Li and Xinshuo Weng and Zhiyu Huang and Zetong Yang and Hongyang Li and Igor Gilitschenski and Boris Ivanovic and Marco Pavone and Andreas Geiger and Kashyap Chitta},
  title   = {NAVSIM: Data-Driven Non-Reactive Autonomous Vehicle Simulation and Benchmarking},
  journal = {NeurIPS},
  year    = {2024}
}

@article{navsimv2,
  author  = {Wei Cao and Marcel Hallgarten and Tianyu Li and Daniel Dauner and Xunjiang Gu and Caojun Wang and Yakov Miron and Marco Aiello and Hongyang Li and Igor Gilitschenski and Boris Ivanovic and Marco Pavone and Andreas Geiger and Kashyap Chitta},
  title   = {Pseudo-Simulation for Autonomous Driving},
  journal = {CoRL},
  year    = {2025}
}

@article{nuplan,
  author  = {Karnchanachari, Napat and Geromichalos, Dimitris and Tan, Kok Seang and Li, Nanxiang and Eriksen, Christopher and Yaghoubi, Shakiba and Mehdipour, Noushin and Bernasconi, Gianmarco and Fong, Whye Kit and Guo, Yiluan and Caesar, Holger},
  title   = {Towards learning-based planning: The nuPlan benchmark for real-world autonomous driving},
  journal = {ICRA},
  year    = {2024}
}

@article{pdm,
  author  = {Daniel Dauner and Marcel Hallgarten and Andreas Geiger and Kashyap Chitta},
  title   = {Parting with Misconceptions about Learning-based Vehicle Motion Planning},
  journal = {CoRL},
  year    = {2023}
}

@misc{openscene,
  author       = {{OpenScene Contributors}},
  title        = {OpenScene: The Largest Up-to-date 3D Occupancy Prediction Benchmark in Autonomous Driving},
  howpublished = {\url{https://github.com/OpenDriveLab/OpenScene}},
  year         = {2023}
}

@article{internvl3,
  author = {Jinguo Zhu and Weiyun Wang and Zhe Chen and Zhaoyang Liu and Shenglong Ye and Lixin Gu and Hao Tian and Yuchen Duan and Weijie Su and Jie Shao and Zhangwei Gao and Erfei Cui and Xuehui Wang and Yue Cao and Yangzhou Liu and Xingguang Wei and Hongjie Zhang and Haomin Wang and Weiye Xu and Hao Li and Jiahao Wang and Nianchen Deng and Songze Li and Yinan He and Tan Jiang and Jiapeng Luo and Yi Wang and Conghui He and Botian Shi and Xingcheng Zhang and Wenqi Shao and Junjun He and Yingtong Xiong and Wenwen Qu and Peng Sun and Penglong Jiao and Han Lv and Lijun Wu and Kaipeng Zhang and Huipeng Deng and Jiaye Ge and Kai Chen and Limin Wang and Min Dou and Lewei Lu and Xizhou Zhu and Tong Lu and Dahua Lin and Yu Qiao and Jifeng Dai and Wenhai Wang},
  title = {InternVL3: Exploring Advanced Training and Test-Time Recipes for Open-Source Multimodal Models},
  journal = {arXiv:2504.10479},
  year = {2025}
}

@article{qformer,
  author  = {Junnan Li and Dongxu Li and Silvio Savarese and Steven Hoi},
  title   = {BLIP-2: Bootstrapping Language-Image Pre-training with Frozen Image Encoders and Large Language Models},
  journal = {ICML},
  year    = {2023}
}

@article{dinov2,
  author  = {Maxime Oquab and Timoth{\'e}e Darcet and Th{\'e}o Moutakanni and Huy Vo and Marc Szafraniec and Vasil Khalidov and Pierre Fernandez and Daniel Haziza and Francisco Massa and Alaaeldin El-Nouby and Mahmoud Assran and Nicolas Ballas and Wojciech Galuba and Russell Howes and Po-Yao Huang and Shang-Wen Li and Ishan Misra and Michael Rabbat and Vasu Sharma and Gabriel Synnaeve and Hu Xu and Herv{\'e} J{\'e}gou and Julien Mairal and Patrick Labatut and Armand Joulin and Piotr Bojanowski},
  title   = {DINOv2: Learning Robust Visual Features without Supervision},
  journal = {ICLR},
  year    = {2024}
}

@article{lora,
  author  = {Edward J. Hu and Yelong Shen and Phillip Wallis and Zeyuan Allen-Zhu and Yuanzhi Li and Shean Wang and Lu Wang and Weizhu Chen},
  title   = {LoRA: Low-Rank Adaptation of Large Language Models},
  journal = {ICLR},
  year    = {2022}
}

@article{diffusiones,
  author  = {Ziqin Yang and Zhiqi Wang and Xiaosong Chen and others},
  title   = {Diffusion-ES: Gradient-free Planning with Diffusion Models},
  journal = {CVPR},
  year    = {2024}
}

@article{botvinick2001conflict,
  title={Conflict monitoring and cognitive control.},
  author={Botvinick, Matthew M and Braver, Todd S and Barch, Deanna M and Carter, Cameron S and Cohen, Jonathan D},
  journal={Psychological review},
  volume={108},
  number={3},
  pages={624},
  year={2001},
  publisher={American Psychological Association}
}

@article{brown2005learned,
  title={Learned predictions of error likelihood in the anterior cingulate cortex},
  author={Brown, Joshua W and Braver, Todd S},
  journal={Science},
  volume={307},
  number={5712},
  pages={1118--1121},
  year={2005},
  publisher={American Association for the Advancement of Science}
}

@article{memoryagent,
  author       = {Yuyang Hu and Shichun Liu and Yanwei Yue and Guibin Zhang and Boyang Liu and Fangyi Zhu and Jiahang Lin and Honglin Guo and Shihan Dou and Zhiheng Xi and Senjie Jin and Jiejun Tan and Yanbin Yin and Jiongnan Liu and Zeyu Zhang and Zhongxiang Sun and Yutao Zhu and Hao Sun and Boci Peng and Zhenrong Cheng and Xuanbo Fan and Jiaxin Guo and Xinlei Yu and Zhenhong Zhou and Zewen Hu and Jiahao Huo and Junhao Wang and Yuwei Niu and Yu Wang and Zhenfei Yin and Xiaobin Hu and Yue Liao and Qiankun Li and Kun Wang and Wangchunshu Zhou and Yixin Liu and Dawei Cheng and Qi Zhang and Tao Gui and Shirui Pan and Yan Zhang and Philip Torr and Zhicheng Dou and Ji{-}Rong Wen and Xuanjing Huang and Yu{-}Gang Jiang and Shuicheng Yan},
  title        = {Memory in the Age of {AI} Agents},
  journal      = {CoRR},
  volume       = {abs/2512.13564},
  year         = {2025},
  url          = {https://doi.org/10.48550/arXiv.2512.13564},
  doi          = {10.48550/ARXIV.2512.13564},
  eprinttype    = {arXiv},
  eprint       = {2512.13564}
}

@article{memoryaiagents,
  author = {Zihan Wang and Qianqian Chen and Zhaopeng Ma and others},
  title = {The Survey: Memory Management in Large Language Models},
  journal = {arXiv:2405.05579},
  year = {2024}
}

@article{transfuser,
  author  = {Kashyap Chitta and Aditya Prakash and Bernhard Jaeger and Zehao Yu and Katrin Renz and Andreas Geiger},
  title   = {TransFuser: Imitation with Transformer-Based Sensor Fusion for Autonomous Driving},
  journal = {IEEE Transactions on Pattern Analysis and Machine Intelligence},
  volume  = {45},
  number  = {11},
  pages   = {12878--12895},
  year    = {2023},
  doi     = {10.1109/TPAMI.2022.3200245}
}

@article{yao2025drivesuprim,
  author  = {Jiahao Yao and others},
  title   = {DriveSuprim: A Robust End-to-End Autonomous Driving Framework},
  journal = {AAAI},
  year    = {2026}
}

@article{gtrs,
  author = {Zhenxin Li and Wenhao Yao and Zi Wang and Xinglong Sun and Joshua Chen and Nadine Chang and Maying Shen and Zuxuan Wu and Shiyi Lan and Jose M. Alvarez},
  title = {Generalized Trajectory Scoring for End-to-end Multimodal Planning},
  journal = {arXiv:2506.06664},
  year = {2025}
}

@article{shao2024lmdrive,
  author  = {Shao, Hao and Hu, Yuxuan and Wang, Letian and Song, Guanglu and Waslander, Steven L and Liu, Yu and Li, Hongsheng},
  title   = {Lmdrive: Closed-loop end-to-end driving with large language models},
  journal = {CVPR},
  year    = {2024}
}

@article{dreamerad,
  author  = {Pengxuan Yang and Yupeng Zheng and Deheng Qian and Zebin Xing and Qichao Zhang and Linbo Wang and Yichen Zhang and Shaoyu Guo and Zhongpu Xia and Qiang Chen and Junyu Han and Lingyun Xu and Yifeng Pan and Dongbin Zhao},
  title   = {DreamerAD: Efficient Reinforcement Learning via Latent World Model for Autonomous Driving},
  journal = {ECCV},
  year    = {2026}
}

@article{planagent,
  author  = {Yupeng Zheng and Zebin Xing and Qichao Zhang and Bu Jin and Pengfei Li and Yuhang Zheng and Zhongpu Xia and Kun Zhan and Xianpeng Lang and Yaran Chen and Dongbin Zhao},
  title   = {PlanAgent: A Multi-modal Large Language Agent for Closed-loop Vehicle Motion Planning},
  journal = {IEEE Transactions on Cognitive and Developmental Systems},
  year    = {2026},
  note    = {Early Access},
  doi     = {10.1109/TCDS.2026.3664120}
}

@article{meanfuser,
  author  = {Junli Wang and Yinan Zheng and Xueyi Liu and Zebin Xing and Pengfei Li and Guang Li and Kun Ma and Guang Chen and Hangjun Ye and Zhongpu Xia and Long Chen and Qichao Zhang},
  title   = {MeanFuser: Fast One-Step Multi-Modal Trajectory Generation and Adaptive Reconstruction via MeanFlow for End-to-End Autonomous Driving},
  journal = {CVPR},
  year    = {2026}
}

@article{beyondimitation,
  author  = {Junli Wang and Zhihua Hua and Xueyi Liu and Zebin Xing and Haochen Tian and Kun Ma and Hangjun Ye and Guang Chen and Long Chen and Qichao Zhang},
  title   = {Beyond Imitation: Learning Safe End-to-End Autonomous Driving from Hard Negatives},
  journal = {ECCV},
  year    = {2026}
}

@article{datascaling,
  author  = {Yupeng Zheng and Pengxuan Yang and Zhongpu Xia and Qichao Zhang and Yuhang Zheng and Songen Gu and Bu Jin and Teng Zhang and Ben Lu and Chao Han and Xianpeng Lang and Dongbin Zhao},
  title   = {Data Scaling Laws for Imitation Learning-Based End-to-End Autonomous Driving},
  journal = {ICRA},
  year    = {2026}
}

@article{soad,
  author  = {Yinfeng Gao and Deqing Liu and Yupeng Zheng and Qichao Zhang and Da-Wei Ding and Dongbin Zhao},
  title   = {SoAD: Safety-Oriented Value Estimation for Enhanced Closed-Loop End-to-End Autonomous Driving},
  journal = {IEEE Transactions on Systems, Man, and Cybernetics: Systems},
  volume  = {56},
  number  = {8},
  pages   = {4942--4955},
  year    = {2026},
  doi     = {10.1109/TSMC.2026.3688954}
}

@article{d2skill,
  author  = {Songjun Tu and Chengdong Xu and Qichao Zhang and Yaocheng Zhang and Xiangyuan Lan and Linjing Li and Dong Li and Dongbin Zhao},
  title   = {Dynamic Dual-Granularity Skill Bank for Agentic RL},
  journal = {CoRR},
  volume  = {abs/2603.28716},
  year    = {2026},
  eprint  = {2603.28716},
  archivePrefix = {arXiv},
  doi     = {10.48550/arXiv.2603.28716}
}

@article{memchain,
  author  = {Yiwen Ma and Songjun Tu and Qichao Zhang and Dong Li and Linjing Li and Dongbin Zhao},
  title   = {MemChain: Learning Interpretable Memory Traces for Memory-Augmented LLM Agents},
  journal = {CoRR},
  volume  = {abs/2607.24079},
  year    = {2026},
  eprint  = {2607.24079},
  archivePrefix = {arXiv},
  doi     = {10.48550/arXiv.2607.24079}
}

@article{ucob,
  author  = {Songjun Tu and Chengdong Xu and Qichao Zhang and Yiwen Ma and Yaocheng Zhang and Linjing Li and Dong Li and Xiangyuan Lan and Dongbin Zhao},
  title   = {UCOB: Learning to Utilize and Evolve Agentic Skills via Credit-Aware On-Policy Bidirectional Self-Distillation},
  journal = {CoRR},
  volume  = {abs/2606.29502},
  year    = {2026},
  eprint  = {2606.29502},
  archivePrefix = {arXiv},
  doi     = {10.48550/arXiv.2606.29502}
}

@article{xing2025trajmoe,
  author        = {Zebin Xing and Pengxuan Yang and Linbo Wang and Yichen Zhang and Yiming Hu and Yupeng Zheng and Junli Wang and Yinfeng Gao and Guang Li and Kun Ma and Long Chen and Zhongpu Xia and Qichao Zhang and Hangjun Ye and Dongbin Zhao},
  title         = {{TrajMoE}: Scene-Adaptive Trajectory Planning with Mixture of Experts and Reinforcement Learning},
  journal       = {arXiv preprint arXiv:2512.07135},
  year          = {2025},
  eprint        = {2512.07135},
  archivePrefix = {arXiv},
  doi           = {10.48550/arXiv.2512.07135},
  url           = {https://arxiv.org/abs/2512.07135}
}

\clearpage
\appendix
\setcounter{page}{1}
\section{Training Details of Base Model}

The training process consists of two stages: VLM pre-training and action decoder optimization. Note that the Compress Module is not included in the VLM pre-training stage; instead, it is trained jointly with the Action Decoder.

\textbf{VLM Pre-training.} We initialize the VLM backbone (excluding the Compress Module) with InternVL3 and perform domain-specific pre-training on large-scale autonomous driving question-answering data. Specifically, we adopt the dataset collected by RecogDrive, which aggregates 12 public datasets including Talk2Car, SUTD, NuScenes-QA, OmniDrive, totaling 3.1 million question-answer pairs covering perception, prediction, and planning tasks. After filtering for high-quality samples, approximately 775K QA pairs are retained for NAVSIM training. Through this process, the VLM acquires specialized reasoning capabilities for autonomous driving scenarios, providing effective decision guidance for the downstream Action Decoder.

\textbf{Action Decoder Training.} After VLM pre-training, we freeze the VLM parameters and focus on training the Action Decoder, which consists of two components: the Trajectory Head for generating diverse trajectory proposals and the Score Head for evaluating and selecting the optimal trajectory. Following a similar training paradigm to iPad, we jointly optimize both heads with a combined loss function:

\begin{equation}
    \mathcal{L}_{\text{total}} = \mathcal{L}_{\text{traj}} + \mathcal{L}_{\text{score}},
\end{equation}

where the trajectory loss $\mathcal{L}_{\text{traj}}$ supervises the generation of candidate trajectories, and the score loss $\mathcal{L}_{\text{score}}$ trains the Score Head to predict the quality of each proposal.

For the Trajectory Head, given $N$ generated proposals $\{\hat{\tau}_i\}_{i=1}^N$ and the ground-truth human trajectory $\tau^*$, we compute the trajectory loss using the minimum L1 distance over all proposals:
\begin{equation}
    \mathcal{L}_{\text{traj}} = \min_{i \in \{1, \dots, N\}} \|\hat{\tau}_i - \tau^*\|_1.
\end{equation}
This min-over-$N$ formulation encourages diverse trajectory generation while ensuring at least one proposal closely matches the expert demonstration.

For the Score Head, we train the model to predict the quality of each trajectory proposal using binary cross-entropy loss. Specifically, each proposal $\hat{\tau}_i$ is evaluated by the PDM scorer to obtain its quality score $s_i \in [0, 1]$, which serves as the soft target. The Score Head predicts a score $\hat{s}_i$ for each proposal, and the score loss is computed as:
\begin{equation}
    \mathcal{L}_{\text{score}} = -\frac{1}{N} \sum_{i=1}^{N} \left[ s_i \log(\hat{s}_i) + (1 - s_i) \log(1 - \hat{s}_i) \right].
\end{equation}

\section{Implementation Details}
All training and inference are conducted on NVIDIA H20 GPUs. For base model training, we use a learning rate of $1\times10^{-4}$, AdamW optimizer, and distribute training across 16 H20 GPUs.

\textbf{Hyperparameters.} For the Retrieve Model, we set $\alpha=10$ to balance the agent and map embedding losses, as the agent loss magnitude is considerably smaller than the map loss. During memory generation, we use $\beta=0.5$ to identify failure cases, following the NAVSIM scoring convention where NC and DAC below 0.5 indicate collisions or drivable area violations. The latent similarity threshold is set to $\lambda=0.9$. For test-time training TTT, we adopt AdamW with learning rate $2\times10^{-4}$ and 3 optimization steps. The memory size is limited to approximately 4K entries for the Base configuration and 10K for the Scale configuration.

\textbf{Deployment Optimizations.} We implement several engineering improvements for efficient deployment. During inference with TTT, we observe that NAVSIM contains numerous empty road segments with no surrounding vehicles or pedestrians. Direct retrieval using agent embeddings alone introduces noise in these scenarios. We therefore employ a hierarchical retrieval strategy: first filtering candidates using map embeddings to obtain top-$k_1$ scenes, then refining with agent embeddings to select top-$k_2$ matches, where $k_1 = 3k_2$, $k_2 = 3$ .

\section{Definition of PDMS and EPDMS}

For completeness, we summarize the evaluation metrics used in NAVSIMv1 and NAVSIMv2, both based on the PDM family of scores.

\textbf{PDMS on NAVSIMv1.} The PDMS combines several sub-metrics via multiplicative penalties and weighted additive terms.
Specifically, it includes two multiplier sub-metrics and three weighted sub-metrics:
No at-fault Collisions (NC) penalizes ego-responsible collisions, taking values in $\{0, \tfrac{1}{2}, 1\}$;
Drivable Area Compliance (DAC) checks whether the ego vehicle stays within drivable regions, taking values in $\{0, 1\}$;
Ego Progress (EP) measures the normalized longitudinal progress along the planned route, with a continuous range $[0,1]$ and weight $w_{\text{EP}}=5$;
Time to Collision (TTC) captures safety margins with respect to surrounding agents, with a binary outcome $\{0,1\}$ and weight $w_{\text{TTC}}=5$;
Comfort (C) evaluates ride quality by checking acceleration and jerk limits, with a binary outcome $\{0,1\}$ and weight $w_{\text{C}}=2$.
PDMS is computed as:
\begin{equation}
  \text{PDMS} = \!\left(\prod_{m\in\{\text{NC}, \text{DAC}\}} \!m(\text{agent})\right) \cdot \left(\frac{\sum_{m\in\{\text{TTC}, \text{EP}, \text{C}\}} w_m \cdot m(\text{agent})}{\sum_{m\in\{\text{TTC}, \text{EP}, \text{C}\}} w_m}\right).
\end{equation}

\textbf{EPDMS on NAVSIMv2.} NAVSIMv2 extends PDMS into the EPDMS by introducing two new multiplier metrics, two new weighted metrics, and a false-positive penalty filtering mechanism.
The full composition of the EPDMS is summarized in Table~\ref{tab:epdms_metrics}.

\begin{table}[h]
\centering
\caption{Sub-metrics of EPDMS. Multiplier metrics are safety-critical and directly scale the final score.}
\label{tab:epdms_metrics}
\resizebox{\columnwidth}{!}{%
\begin{tabular}{lcc}
\toprule
\textbf{Metric} & \textbf{Weight} & \textbf{Range} \\
\midrule
No at-fault Collisions (NC) & multiplier & $\{0, \tfrac{1}{2}, 1\}$ \\
Drivable Area Compliance (DAC) & multiplier & $\{0, 1\}$ \\
Driving Direction Compliance (DDC) & multiplier & $\{0, \tfrac{1}{2}, 1\}$ \\
Traffic Light Compliance (TLC) & multiplier & $\{0, 1\}$ \\
Ego Progress (EP) & 5 & $[0, 1]$ \\
Time to Collision (TTC) & 5 & $\{0, 1\}$ \\
Lane Keeping (LK) & 2 & $\{0, 1\}$ \\
History Comfort (HC) & 2 & $\{0, 1\}$ \\
Extended Comfort (EC) & 2 & $\{0, 1\}$ \\
\bottomrule
\end{tabular}%
}
\end{table}

Compared with PDMS, the four newly introduced sub-metrics are:
Driving Direction Compliance (DDC) penalizes wrong-way driving, taking values in $\{0, \tfrac{1}{2}, 1\}$;
Traffic Light Compliance (TLC) penalizes traffic signal violations, taking values in $\{0, 1\}$;
Lane Keeping (LK) penalizes driving too far from the centerline for an extended time (disabled at intersections where centerline annotations may be unreliable), with a binary outcome $\{0,1\}$ and weight $w_{\text{LK}}=2$;
Extended Comfort (EC) compares trajectory outputs of subsequent frames and their resulting dynamic states, penalizing discrepancies in acceleration and jerk between frames, with a binary outcome $\{0,1\}$ and weight $w_{\text{EC}}=2$.
In addition, the original Comfort (C) is improved to History Comfort (HC), which also evaluates how the planned trajectory matches the vehicle's motion history.

To reduce false-positive penalties, EPDMS introduces a filtering function that disables penalties when the human driver also commits the same violation:
\begin{equation}
  \text{filter}_m(\text{agent}, \text{human}) =
  \begin{cases}
    1.0, & \text{if } m(\text{human}) = 0, \\
    m(\text{agent}), & \text{otherwise.}
  \end{cases}
\end{equation}
The full EPDMS is then computed as:
\begin{equation}
  \text{EPDMS} = \!\left(\prod_{m\in\mathcal{M}_{\text{mul}}} \!\text{filter}_m\right) \cdot \left(\frac{\sum_{m\in\mathcal{M}_{\text{add}}} w_m \cdot \text{filter}_m}{\sum_{m\in\mathcal{M}_{\text{add}}} w_m}\right),
\end{equation}

NAVSIMv2 evaluates planners in two stages. Stage~1: The planner generates a trajectory from the original scene; sub-metrics are computed over a non-reactive rollout. Stage~2: Pre-computed follow-up scenes (representing continuations such as near-miss situations) are evaluated to assess robustness under extended horizons. EPDMS aggregates all sub-metrics over both stages; failures in either stage are strongly penalized, making EPDMS more stringent and closed-loop–aligned than PDMS. We report individual sub-metrics (NC, DAC, EP, TTC, C, LK, DDC, TLC, EC) with final PDMS (NAVSIMv1) and EPDMS (NAVSIMv2) per the official protocol.

\section{Additional Ablation Study}
In this section, we conduct further ablation experiments on the learning rate and gradient steps during TTT.

\textbf{Ablation of Params in TTT.}
Table~\ref{tab:ablation_ttt_hp} ablates learning rate and gradient steps for the TTT optimization using the AdamW optimizer. DriveVLA-M0 demonstrates strong robustness across a wide hyperparameter range: PDMS varies by at most $0.3$ across all combinations tested.

Stable performance under step counts of 1 to 5 and learning rates of $2\times10^{-5}$ to $5\times10^{-5}$ reflects the inherent regularization of low-rank adaptation, which constrains parameter updates to a small subspace and prevents catastrophic forgetting. Notably, even with only 1 gradient step, the model achieves competitive performance (92.0--92.2 PDMS), enabling rapid adaptation for latency-critical scenarios. For this paper, we adopt learning rate $2\times10^{-4}$ and 3 steps as the default configuration.

\begin{table}[h]
\centering
\caption{Robustness ablation on TTT learning rate and gradient steps.}
\label{tab:ablation_ttt_hp}
\begin{tabular}{cccccccc}
\toprule
LR & Steps & PDMS & NC & DAC & EP & TTC & C \\
\midrule
$5\times10^{-5}$ & 1  & 92.2 & 98.8 & 97.8 & 89.6 & 94.9 & 99.9 \\
$1\times10^{-4}$ & 1  & 92.2 & 98.9 & 97.8 & 89.6 & 94.9 & 99.9 \\
$2\times10^{-4}$ & 1  & 92.0 & 98.9 & 97.6 & 89.4 & 94.9 & 99.9 \\
\midrule
$5\times10^{-5}$ & 3  & \textbf{92.3} & 98.9 & 97.9 & \textbf{89.7} & 94.9 & 99.9 \\
$1\times10^{-4}$ & 3  & 92.3 & 98.9 & 97.8 & 89.6 & 95.1 & 99.9 \\
$2\times10^{-4}$ & 3  & 92.3 & \textbf{99.0} & 97.7 & 89.5 & 95.2 & 99.9 \\
\midrule
$5\times10^{-5}$ & 5  & 92.3 & \textbf{99.0} & 97.8 & 89.6 & \textbf{95.0} & 99.9 \\
$1\times10^{-4}$ & 5  & 92.2 & \textbf{99.0} & 97.7 & 89.5 & \textbf{95.0} & 99.9 \\
$2\times10^{-4}$ & 5  & 92.1 & \textbf{99.0} & 97.6 & 89.3 & 95.2 & 99.9 \\
\bottomrule
\end{tabular}
\end{table}

\section{Additional Visualization}
\begin{figure*}[!h]
  \centering
  \includegraphics[width=1.0\textwidth]{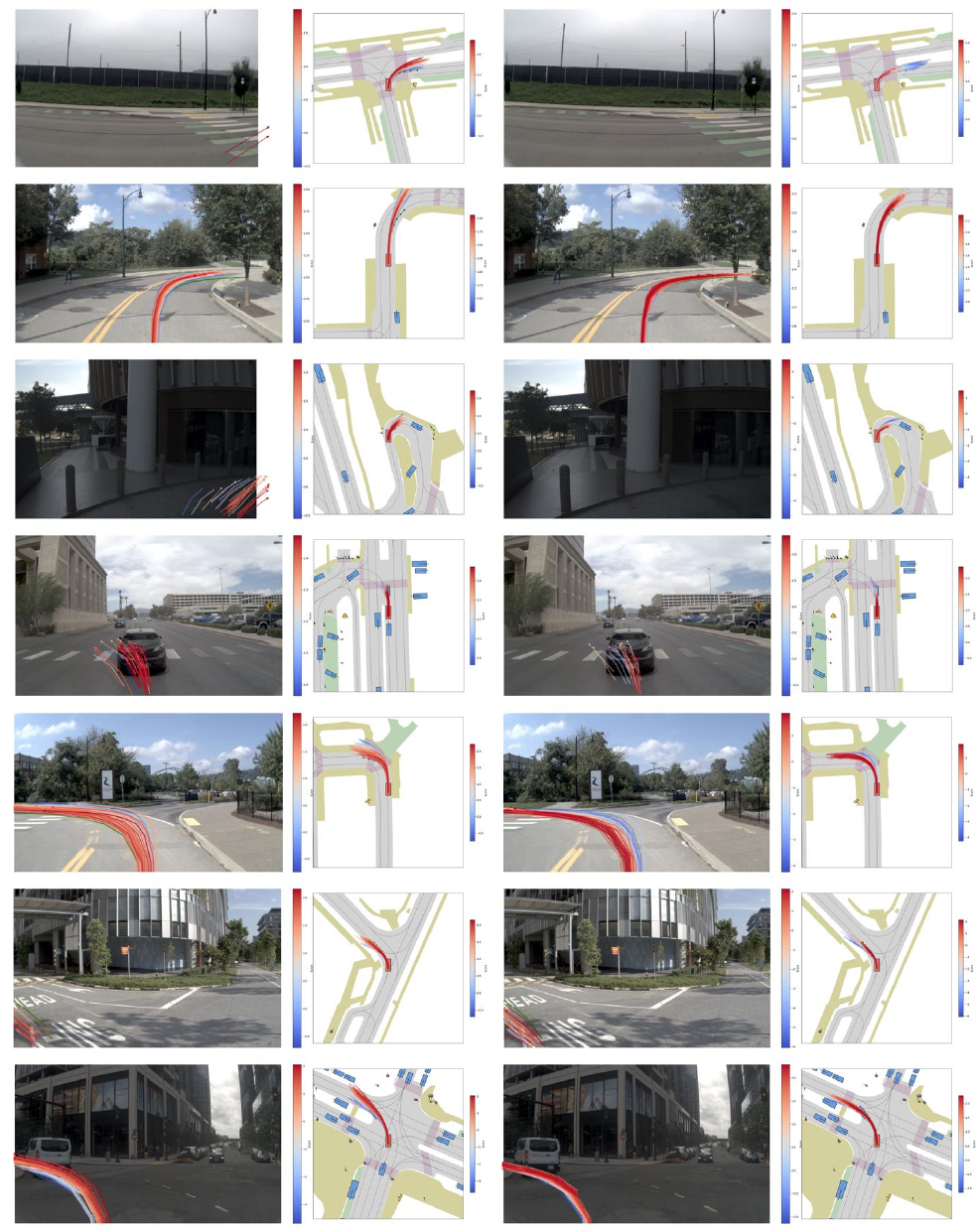}
  \caption{Comparison of trajectory clusters before and after memory injection. In each pair, the left shows trajectories generated by the base model, and the right shows trajectories after injection.}
  \label{fig:comparsion_base_drivevla}
\end{figure*}

\begin{figure*}[!h]
  \centering
  \includegraphics[width=\textwidth]{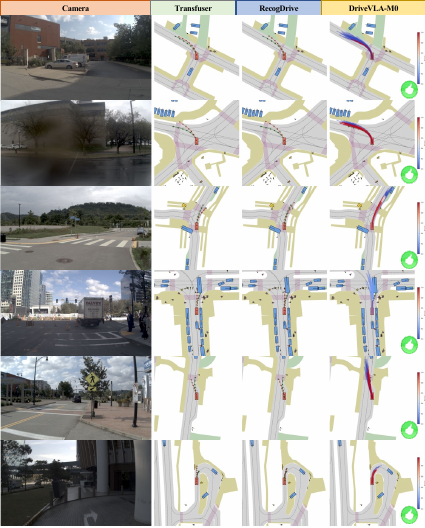}
  \caption{Trajectory comparison with prior methods. We select Transfuser as a representative end-to-end method and RecogDrive as a representative VLM-based method, and compare their generated trajectories with ours.}
  \label{fig:comparison_methods}
\end{figure*}

\begin{figure*}[!h]
  \centering
  \includegraphics[width=\textwidth]{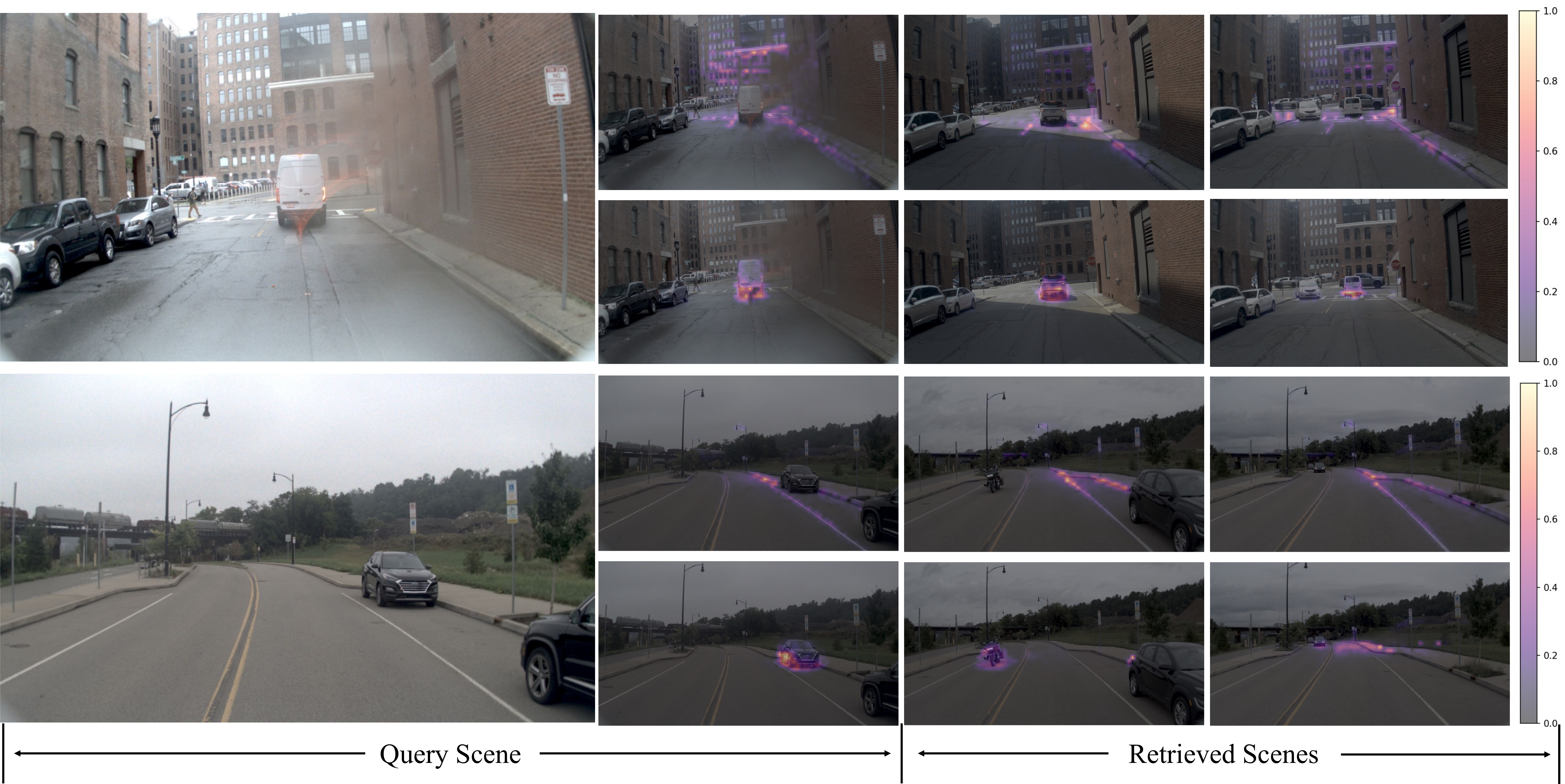}
  \caption{Attention maps of query and retrieved scenes. In each group, the top row shows the map embedding attention and the bottom row shows the agent embedding attention. Yellow indicates higher attention, and gray-purple indicates lower attention.}
  \label{fig:appendix_retrieve_demo}
\end{figure*}

We provide additional qualitative results to complement the quantitative analysis. Figure~\ref{fig:comparsion_base_drivevla} visualizes the effect of memory injection by contrasting trajectory clusters before and after TTT. Figure~\ref{fig:comparison_methods} presents a trajectory-level comparison with representative prior methods. Finally, Figure~\ref{fig:appendix_retrieve_demo} shows attention maps of the retrieval module, revealing which regions the model attends to when matching query and retrieved scenes.

\end{document}